\documentclass[11pt,a4paper]{article}
\usepackage[hyperref]{emnlp2020}
\usepackage{times}
\usepackage{latexsym}

\usepackage{microtype}

\aclfinalcopy % Uncomment this line for the final submission
\def\aclpaperid{1012} %  Enter the acl Paper ID here

\usepackage{xcolor}
\usepackage{url}
\usepackage{booktabs}       % professional-quality tables
\usepackage{amsfonts}       % blackboard math symbols
\usepackage{nicefrac}       % compact symbols for 1/2, etc.
\usepackage{graphicx}
\usepackage{subfigure}
\usepackage{pifont}
\usepackage{url}
\usepackage[utf8]{inputenc} % allow utf-8 input
\usepackage[T1]{fontenc}    % use 8-bit T1 fonts
\usepackage{url}            % simple URL typesetting
\usepackage{xcolor}
\usepackage{soul}
\usepackage[utf8]{inputenc}
\usepackage{amsmath}
\usepackage{multirow}
\usepackage{float}
\usepackage{epsfig}

\title{UniConv: A Unified Conversational Neural Architecture for Multi-domain Task-oriented Dialogues}

\author{Hung Le$^{\dag}{^\S}$, Doyen Sahoo$^{\ddag}$, Chenghao Liu$^{\dag}$, Nancy F. Chen$^{\S}$, Steven C.H. Hoi$^{\dag\ddag}$ \\
  $^{\dag}$ Singapore Management University \\
  \texttt{\{hungle.2018,chliu\}@smu.edu.sg} \\
  $^{\ddag}$ Salesforce Research Asia\\
  \texttt{\{dsahoo,shoi\}@salesforce.com} \\ 
  $^\S$Institute for Infocomm Research, A*STAR \\ 
  \texttt{nfychen@i2r.a-star.edu.sg}}
  
\date{}

\begin{document}
\maketitle
\begin{abstract}
Building an end-to-end conversational agent for multi-domain task-oriented dialogues has been an open challenge for two main reasons. First, tracking dialogue states of multiple domains is non-trivial as the dialogue agent must obtain complete states from all relevant domains, some of which might have shared slots among domains as well as unique slots specifically for one domain only. Second, the dialogue agent must also process various types of information across domains, including dialogue context, dialogue states, and database, to generate natural responses to users. Unlike the existing approaches that are often designed to train each module separately, we propose ``UniConv" --- a novel unified neural architecture for end-to-end conversational systems in multi-domain task-oriented dialogues, which is designed to jointly train (i) a Bi-level State Tracker which tracks dialogue states by learning signals at both slot and domain level independently, and (ii) a Joint Dialogue Act and Response Generator which incorporates information from various input components and models dialogue acts and target responses simultaneously. We conduct comprehensive experiments in dialogue state tracking, context-to-text, and end-to-end settings on the MultiWOZ2.1 benchmark, achieving superior performance over competitive baselines.  
\end{abstract}

\section{Introduction}
% \usepackage{multirow}
\begin{table*}[]
\resizebox{1.0\textwidth}{!} {
\begin{tabular}{ll}
\hline 
%\multicolumn{1}{c}{\textbf{Turn}} & \multicolumn{1}{c}{\textbf{User utterance}}                                                                                                                                                                                                             & \multicolumn{1}{c}{\textbf{System response}}                                                                                                   \\\hline 
\textbf{Human}: hello                                                                                                                                                                                                                                                   & \textbf{Dialogue agent}: hi ! how can i help you ?                                                                                                                      \\
                                  \textit{Dialogue state: \{\}}                                                                                                                                                                                                                           & \textit{Dialogue acts: \textcolor{blue}{{[}greeting{]}}}                                                                                                         \\\hline 
\textbf{Human}: i would like a place to eat in the expensive price range                                                                                                                                                                                                & \textbf{Dialogue agent}: sure , what type of food are you interested in ?                                                                                               \\
                                  \textit{Dialogue state: \{\textcolor{red}{restaurant: \{pricerange: expensive\}}\}}                                                                                                                                                                                      & \textit{Dialogue acts: \textcolor{blue}{{[}request-food{]}}}                                                                                                     \\\hline 
\textbf{Human}: could you make a suggestion ? one in the centre ?                                                                                                                                                                                                       & \begin{tabular}[c]{@{}l@{}}\textbf{Dialogue agent}: fitzbillies restaurant is an expensive british  \\ restaurant in the centre . can i book that for you ?\end{tabular} \\
                                  \textit{Dialogue state: \{\textcolor{red}{restaurant: \{pricerange: expensive, area: centre\}}\}}                                                                                                                                                                        & \textit{Dialogue acts: \textcolor{blue}{{[}inform-restaurant, request-booking{]}}}                                                                               \\ 
...                                                                                                                                                                                                                                                     & ...                                                                                                                                            \\\hline 
\textbf{Human}: also , i need the number for kings hedges learner pool .                                                                                                                                                                                                & \begin{tabular}[c]{@{}l@{}}\textbf{Dialogue agent}: the phone number for the pool is 01234567 , \\is there something else i can help you ?\end{tabular}                \\
\textit{\begin{tabular}[c]{@{}l@{}}Dialogue state: \{\textcolor{red}{restaurant: \{pricerange: expensive, area: centre,}\\ \textcolor{red}{name=fizbillies restaurant, request={[}address{]}}\},  \textcolor{orange}{attraction:} \\ \textcolor{orange}{\{name: kings hedges learner pool, request={[}phone{]}\}}\}\end{tabular}} & \textit{Dialogue acts: \textcolor{blue}{{[}inform-phone{]}}}        \\
                                  \hline 
\end{tabular}
}
\caption{Example of a multi-domain dialogue with two domains: \emph{restaurant} and \emph{attraction}.}
%Each row represents a dialogue turn with annotation of dialogue state and dialogue acts.}
\label{fig:data}
%\vspace{-0.6cm}
\end{table*}

%\begin{figure}[t]
%	\includegraphics[width=1.0\columnwidth]{sample_dial_v4.pdf}
%	\centering
%	\caption{Examples of multi-domain dialogues. \textit{$U$}: user utterance, \textit{$S$}: dialogue state, \textit{$A$}: dialogue act, \textit{$R$}: system response. The subscript indicates dialogue turn.}
%\vspace{-0.1in}
%%	\label{fig:data}
%	\vspace{-0.2in}
%\end{figure}

A conventional approach to task-oriented dialogues is to solve four distinct tasks: (1) natural language understanding (NLU) which parses user utterance into a semantic frame, (2) dialogue state tracking (DST) which updates the slots and values from semantic frames to the latest values for knowledge base retrieval, (3) dialogue policy which determines an appropriate dialogue act for the next system response, and (4) response generation which generates a natural language sequence conditioned on the dialogue act. 
This traditional pipeline modular framework has achieved remarkable successes in task-oriented dialogues \cite{wen2016network, liu2017end, williams-etal-2017-hybrid, zhao-etal-2017-generative}. However, such kind of dialogue system is not fully optimized as the modules are loosely integrated and often not trained jointly in an end-to-end manner, and thus may suffer from increasing error propagation between the modules as the complexity of the dialogues evolves. 
%A%n end-to-end task-oriented dialogue system assists humans to complete one or multiple tasks by exchanging domain-specific information with humans through conversations. 
%In a task-oriented dialogue system, the Dialogue State Tracking (DST) module is responsible for updating dialogue states (essentially, what the user wants) at each dialogue turn \cite{henderson2014second}.
%The DST supports the dialogue agent to steer the conversation towards task completion. 

A typical case of a complex dialogue setting is when the dialogue extends over multiple domains. 
%In a multi-domain dialogues, a dialogue state consist of  all \textit{inform} slots - information to query a given knowledge base or database (DB) e.g. \textit{restaurant\_pricerange=expensive}, and all \textit{request} slots - information to be returned to the users e.g. \textit{restaurant\_address}. 
%Developing end-to-end task-oriented dialogue systems has gained interest from the research community \cite{wen2016network, eric2017key, lei2018sequicity, wu2019global}. 
%Task-oriented dialogues can be categorized as either single-domain or multi-domain dialogues. In single-domain dialogues, humans converse with the dialogue agent to complete tasks of one domain. In contrast, in multi-domain dialogues, the tasks of interest can come from different domains. 
A dialogue state in a multi-domain dialogue should include slots of all applicable domains up to the current turn (See Table \ref{fig:data}). 
%Tracking dialogue states in multi-domain dialogues is challenging because of the diverse information from multiple domains. 
Each domain can have shared slots that are common among domains or unique slots that are not shared with any. Directly applying single-domain DST to multi-domain dialogues is not straightforward because the dialogue states extend to multiple domains. A possible approach is to process a dialogue of $N_{D}$ domains multiple times, each time obtaining a dialogue state of one domain. However, this approach does not allow learning co-reference in dialogues in which users can switch from one domain to another.

%Another important components of task-oriented dialogues are Dialogue Policy and Natural Language Generator (NLG). 
%Dialogue Policy decides what \textit{dialogue act} a dialogue agent should follow. Equivalent to user intention, an \textit{act} can be defined as the overall semantic goal which conditions a system response. For example, a ``greeting" act defines that dialogue agent should greet the users, usually at the beginning of a conversational episode. 
%A ``confirm booking" act dictates the dialogue agent to provide an affirmative response with booking information, typically whenever all required \textit{inform} slots are given. 
%Given a specific \textit{act}, NLG constructs a natural response that is coherent and relevant to the users. Multi-domain NLG is difficult because output responses can include domain-specific vocabulary. A successful NLG must incorporate all signals, states from DST and database, and \textit{act} from Dialogue Policy. To generate correct responses, it is critical for dialogue models to obtain dependencies between these signals in fine granularity.%, especially for responses of multiple domains.

As the number of dialogue domains increases, traditional pipeline approaches propagate errors from dialogue states to dialogue policy and subsequently, to natural language generator. 
Recent efforts \cite{eric2017key, madotto2018mem2seq, wu2019global} address this problem with an integrated sequence-to-sequence structure. These approaches often consider knowledge bases as memory tuples rather than relational entity tables. While achieving impressive performance, these approaches are not scalable to large-scale knowledge-bases, e.g. thousands of entities, as the memory cost to query entity attributes increases substantially. Another limitation of these approaches is the absence of dialogue act modelling. Dialogue act is particularly important in task-oriented dialogues as it determines the general decision towards task completion before a dialogue agent can materialize it into natural language response (See Table \ref{fig:data}).  

To tackle the challenges in multi-domain task-oriented dialogues while reducing error propagation among dialogue system modules and keeping the models scalable, we propose UniConv, a unified neural network architecture for end-to-end dialogue systems.
%joint state tracking and response generation with inherent dialogue act modeling. 
UniConv consists of a Bi-level State Tracking (BDST) module which embeds natural language understanding as it can directly parse dialogue context into a structured dialogue state rather than relying on the semantic frame output from an NLU module in each dialogue turn.
BDST implicitly models and integrates slot representations from dialogue contextual cues to directly generate slot values in each turn and thus, remove the need for explicit slot tagging features from an NLU. 
This approach is more practical than the traditional pipeline models as we do not need slot tagging annotation.
Furthermore, BDST tracks dialogue states in dialogue context in both slot and domain levels. The output representations from two levels are combined in a {\it late fusion} approach to learn multi-domain dialogue states. Our dialogue state tracker disentangles slot and domain representation learning while enabling deep learning of shared representations of slots common among domains.
%Compared to existing DST methods, our DST follows a generation-based principle, which generates slot values rather than relying on a known ontology of (slot, value) pairs. The generation-based method is more appropriate to track open-vocabulary slots such as entity names and time but is usually subject to lower accuracy than retrieval-based approaches. However, BDST can generate dialogue states effectively and increases model performance over existing approaches. 

UniConv integrates BDST with a Joint Dialogue Act and Response Generator (DARG) that simultaneously models dialogue acts and generates system responses by learning a latent variable representing dialogue acts and semantically conditioning output response tokens on this latent variable.  
The multi-task setting of DARG allows our models to model dialogue acts and utilize the distributed representations of dialogue acts, rather than hard discrete output values from a dialogue policy module, on output response tokens.
Our response generator incorporates information from dialogue input components and intermediate representations progressively over multiple attention steps. The output representations are refined after each step to obtain high-resolution signals needed to generate appropriate dialogue acts and responses. 
We combine both BDST and DARG for end-to-end neural dialogue systems, from input dialogues to output system responses. 

We evaluate our models on the large-scale MultiWOZ benchmark \citep{budzianowski2018multiwoz}, and compare with the existing methods in DST, context-to-text generation, and end-to-end settings. The promising performance in all tasks validates the efficacy of our method. 
%To the best of our knowledge, we are the first to report results on the end-to-end setting of MultiWOZ benchmark.

\section{Related Work}
\begin{figure*}[htbp]
	%\vspace{-0.2in}
	\centering
	\includegraphics[width=\textwidth]{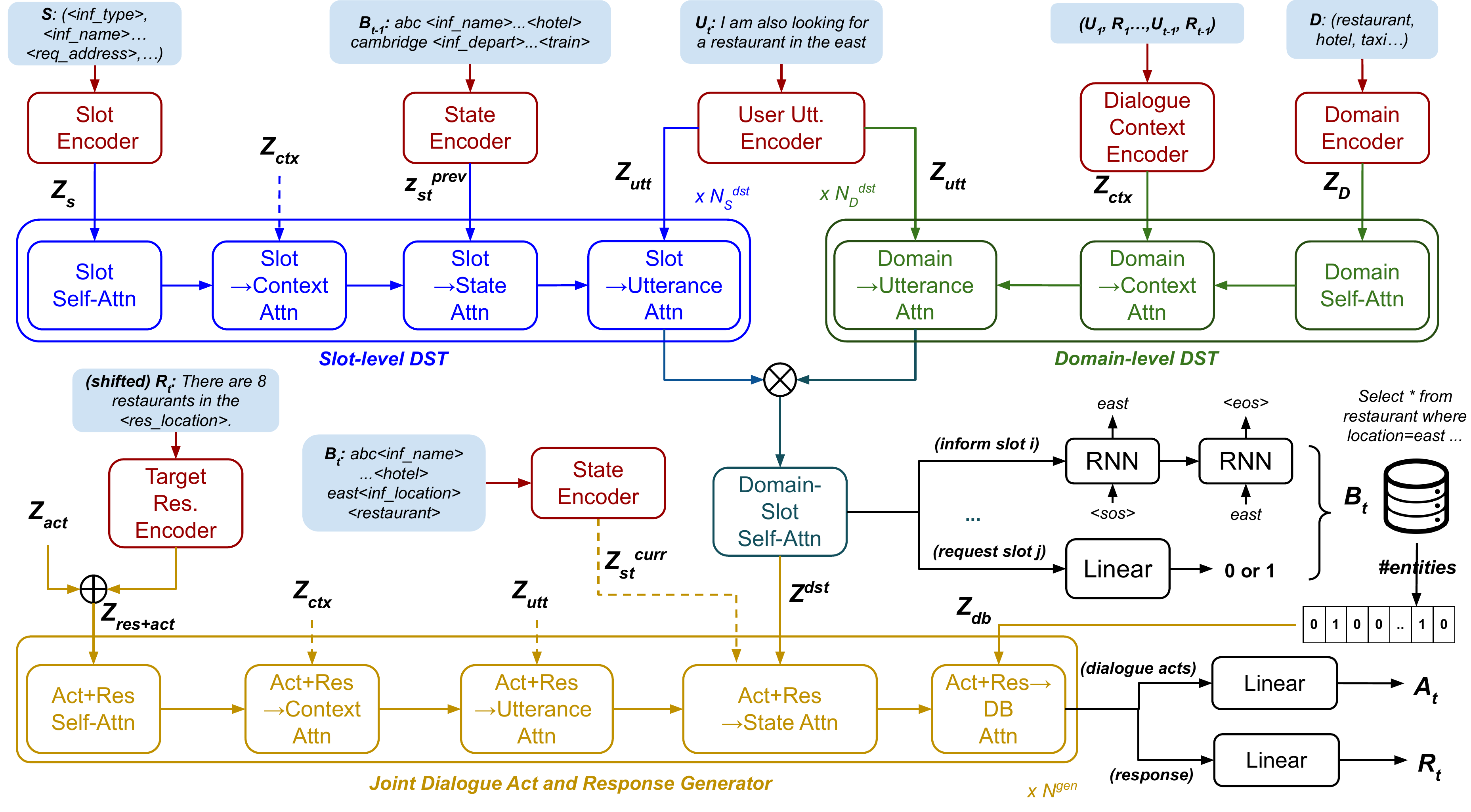}
	%\vspace{-0.2in}
	\caption{Our unified architecture has three components: 
	(1) \textit{Encoders} encode all text input into continuous representations;
    (2) \textit{Bi-level State Tracker (BDST)} includes 2 modules for slot-level and domain-level representation learning; and
    (3) \textit{Joint Dialogue Act and Response Generator (DARG)} obtains dependencies between the target response representations and other dialogue components.}
    %Our response generator is also utilized to model dialogue acts. For simplicity, some layers are not shown, such as Feed Forward, Residual Connection, and Layer Normalization.}
	\label{fig:model}
	%\vspace{-0.2in}
\end{figure*}

\textbf{Dialogue State Tracking}.
Traditionally, DST models are designed to track states of single-domain dialogues such as WOZ \citep{wen2016network} and DSTC2 \citep{henderson2014second} benchmarks.
There have been recent efforts that aim to tackle multi-domain DST such as \citep{ramadan2018large, lee2019sumbt, wu-etal-2019-transferable, goel2019hyst}.
These models can be categorized into two main categories: Fixed vocabulary models \citep{zhong2018global, ramadan2018large, lee2019sumbt}, which assume known slot ontology with a fixed candidate set for each slot. On the other hand, open-vocabulary models \citep{lei2018sequicity, wu-etal-2019-transferable, gao2019dialog, ren-etal-2019-scalable, Le2020Non-Autoregressive} derive the candidate set based on the source sequence i.e. dialogue history, itself.  
 Our approach is more related to the open-vocabulary approach as we aim to generate unique dialogue states depending on the input dialogue.
Different from previous generation-based approaches, 
%our Multi-level State Tracker learns slot and domain representations by incorporating contextual signals . 
our state tracker can incorporate contextual information into domain and slot representations independently.
%and explicitly learns dependencies among slots and domains independently.
%, allowing models to derive relationship between slot pairs and domain pairs independently. 

\textbf{Context-to-Text Generation}.
%\citet{budzianowski2018multiwoz} propose a dialogue task in which given the access to the ground-truth dialogue states, dialogue agent has to construct natural responses to the users. 
This task was traditionally solved by two separate dialogue modules: Dialogue Policy \citep{peng2017composite, peng2018deep} and NLG \citep{wen2016conditional, su-etal-2018-natural}. 
Recent work attempts to combine these two modules to directly generate system responses with or without modeling dialogue acts. 
\citet{zhao-etal-2019-rethinking} models action space of dialogue agent as latent variables.
\citet{chen2019semantically} predicts dialogue acts using a hierarchical graph structure with each path representing a unique act. 
\citet{pei2019modular, peng2019teacher} use multiple dialogue agents, each trained for a specific dialogue domain, and combine them through a common dialogue agent.
\citet{mehri2019structured} models dialogue policy and NLG separately and fuses feature representations at different levels to generate responses. 
 %through a multi-labeled classification task to allow multiple dialogue acts in each turn while semantically conditioning all target tokens on this latent variable in each generation step.
Our models simultaneously learn dialogue acts as a latent variable while allowing semantic conditioning on distributed representations of dialogue acts rather than hard discrete features.

\textbf{End-to-End Dialogue Systems}.
In this task, conventional approaches combine Natural Language Understanding (NLU), DST, Dialogue Policy, and NLG, into a pipeline architecture \citep{wen2016network, bordes2016learning, liu2017end, li2017end, liu-perez-2017-gated, williams-etal-2017-hybrid, zhao-etal-2017-generative, jhunjhunwala-etal-2020-multi}.
Another framework does not explicitly modularize these components but incorporate them through a sequence-to-sequence framework \citep{serban2016building, lei2018sequicity, yavuz-etal-2019-deepcopy} and a memory-based entity dataset of triplets \citep{eric2017copy, eric2017key, madotto2018mem2seq, qin-etal-2019-entity, gangi-reddy-etal-2019-multi, wu2019global}.
These approaches bypass dialogue state and/or act modeling and aim to generate output responses directly. 
They achieve impressive success in generating dialogue responses in open-domain dialogues with unstructured knowledge bases. 
However, in a task-oriented setting with an entity dataset, they might suffer from an explosion of memory size when the number of entities from multiple dialogue domains increases.
Our work is more related to the traditional pipeline strategy
%as we allow our model to explicitly learn dialogue states and acts and enable efficient database search. 
%Especially, database of multi-domain dialogues can contain a large number of entities to be queried.
%Different from pipeline models, 
but we integrate our dialogue models by unifying two major components rather than using the traditional four-module architecture, to alleviate error propagation from upstream to downstream components. 
%Similar to our work, \citet{shu--2019-flexibly} proposes a novel end-to-end architecture for task-oriented dialogue systems. 
Different from prior work such as \cite{shu-etal-2019-flexibly}, our model facilitates multi-domain state tracking and allows learning dialogue acts during response generation. 
%\vspace{-0.2cm}
\section{Method}
%\vspace{-0.2cm}
The input consists of dialogue context of $t-1$ turns, each including a pair of user utterance $U$ and system response $R$, $(U_1, R_1),...,(U_{t-1}, R_{t-1})$, and the user utterance at current turn $U_t$.
A task-oriented dialogue system aims to generate the next response $R_t$.
%, that is not only appropriate to the dialogue context, but also contains the correct information relevant to the current dialogue domains.
The information for responses is typically queried from a database based on the user's provided information i.e. \textit{inform} slots tracked by a DST. 
We assume access to a database of all domains with each column corresponding to a specific slot being tracked. 
We denote the intermediate output, including the dialogue state of current turn $B_t$ and dialogue act as $A_t$. 
We denote the list of all domains $D=(d_1, d_2,...)$, all slots $S=(s_1, s_2, ...)$, and all acts $A=(a_1, a_2, ...)$. 
We also denote the list of all (domain, slot) pairs as $DS=(ds_1, ds_2,...)$.
Note that $\|DS\| \leq \|D\| \times \|S\|$ as some slots might not be applicable in all domains. 
Given the current dialogue turn $t$, we represent each text input as a sequence of tokens, each of which is a unique token index from a vocabulary set $V$: dialogue context $X_\mathrm{ctx}$, current user utterance $X_\mathrm{utt}$, and target system response $X_\mathrm{res}$. 
Similarly, we also represent the list of domains as $X_D$ and the list of slots as $X_S$. 

\noindent In DST, we consider the raw text form of dialogue state of the previous turn $B_{t-1}$, similarly as \cite{lei2018sequicity, budzianowski-vulic-2019-hello}. 
%using the following template: 
%$    \langle\mathrm{value1}\rangle \langle\mathrm{slot1}\rangle
%    \langle\mathrm{value2}\rangle \langle\mathrm{slot2}\rangle... \langle\mathrm{domain1}\rangle  \\
%    \langle\mathrm{value1}\rangle \langle\mathrm{slot1}\rangle
%    \langle\mathrm{value2}\rangle \langle\mathrm{slot2}\rangle... \langle\mathrm{domain2}\rangle ...$
In the context-to-text setting, we assume access to the ground-truth dialogue states of current turn $B_t$.
%, we also consider the raw text form using the same template. 
The dialogue state of the previous and current turn can then be represented as a sequence of tokens $X^\mathrm{prev}_\mathrm{st}$ and $X^\mathrm{curr}_\mathrm{st}$ respectively.
For a fair comparison with current approaches, during inference, we use the model predicted dialogue states $\hat{X}^\mathrm{prev}_\mathrm{st}$ and do not use $X^\mathrm{curr}_\mathrm{st}$ in DST and end-to-end tasks. 
Following \cite{wen-etal-2015-semantically, budzianowski2018multiwoz}, we consider the delexicalized target response $X^\mathrm{dl}_\mathrm{res}$ by replacing tokens of slot values by their corresponding generic tokens to allow learning value-independent parameters.

\noindent Our model consists of 3 major components (See Figure \ref{fig:model}).
First, \textit{Encoders} encode all text input into continuous representations.
To make it consistent, we encode all input with the same embedding dimension. 
Secondly, our \textit{Bi-level State Tracker (BDST)} is used to detect contextual dependencies to generate dialogue states. 
The DST includes 2 modules for slot-level and domain-level representation learning. 
Each module comprises attention layers to project domain or slot representations  and incorporate important information from dialogue context, dialogue state of the previous turn, and current user utterance. 
The outputs 
%of the two modules are combined to create domain-slot joint feature representations. 
are combined as a context-aware vector to decode the corresponding \textit{inform} or \textit{request} slots in each domain. 
Lastly, our \textit{Joint Dialogue Act and Response Generator (DARG)} projects the target system response representations and enhances them with information from various dialogue components. 
Our response generator can also learn a latent representation to generate dialogue acts, which condition all target tokens during each generation step.
\subsection{Encoders}
\label{subsec:enc}
An encoder encodes a text sequence $X$ to a sequence of continuous representation $Z \in \mathbb{R}^{L_X \times d}$. 
$L_X$ is the length of sequence $X$ and $d$ is the embedding dimension. 
Each encoder includes a token-level embedding layer.
The embedding layer is a trainable embedding matrix $E \in \mathbb{R}^{\|V\| \times d}$.
Each row represents a token in the vocabulary set $V$ as a $d$-dimensional vector. 
We denote $E(X)$ as the embedding function that transform the sequence $X$ by looking up the respective token index: $Z_\mathrm{emb}=E(X) \in \mathbb{R}^{L_X \times d}$.
We inject the positional attribute of each token as similarly adopted in \cite{vaswani17attention}. 
The positional encoding is denoted as $PE$. The final embedding is the element-wise summation between token-embedded representations and positional encoded representations with layer normalization \citep{ba2016layer}:
$Z = \mathrm{LayerNorm}(Z_\mathrm{emb} + PE(X)) \in \mathbb{R}^{L_X \times d}$.

The encoder outputs include representations of dialogue context $Z_\mathrm{ctx}$, current user utterance $Z_\mathrm{utt}$, and target response $Z^\mathrm{dl}_\mathrm{res}$.
We also encode the dialogue states of the previous turn and current turn and obtain $Z^\mathrm{prev}_\mathrm{st}$ and $Z^\mathrm{curr}_\mathrm{st}$ respectively.
We encode $X_S$ and $X_D$ using only token-level embedding layer: $Z_S=\mathrm{LayerNorm}(E(X_S))$ and $Z_D=\mathrm{LayerNorm}(E(X_D))$.
During training, we shift the target response by one position to the left side to allow auto-regressive prediction in each generation step. 
We share the embedding matrix $E$ to encode all text tokens except for tokens of target responses as the delexicalized outputs contain different semantics from natural language inputs.

\subsection{Bi-level Dialogue State Tracker (BDST)}
%We propose a multi-level DST whereby domain and slot representations are learned independently.

\textbf{Slot-level DST}.
We adopt the Transformer attention \cite{vaswani17attention}, 
which consists of a dot-product attention with skip connection, to integrate dialogue contextual information into each slot representation. 
We denote $\mathrm{Att}(Z_1, Z_2)$ as the attention operation from $Z_2$ on $Z_1$.
%, including below layers:
%\begin{align}
%    Z^{(1)} &= Z_1 W^{(1)} \in \mathbb{R}^{L_{Z_1} \times d_{\mathrm{att}}} \label{eq1}\\
%    Z^{(2)} &= Z_2 W^{(2)} \in \mathbb{R}^{L_{Z_2} \times d_{\mathrm{att}}}\\
%    S &= \mathrm{Softmax}(Z^{(2)} Z^{(1)}) \in \mathbb{R}^{L_{Z_2} \times L_{Z_1}} \\
%    Z^{(3)} &= \mathrm{ReLU}((S Z_1) W^{(3)}) \in \mathbb{R}^{L_{Z_2} \times d} \\
%    Z^{(4)} &= \mathrm{LayerNorm}(Z_2 + Z^{(3)}) \in \mathbb{R}^{L_{Z_2} \times d} \label{eq5}
%\end{align}
%where 
%$W^{(1)}, W^{(2)}, W^{(3)}$ each has dimensions $\mathbb{R}^{d \times d_{\mathrm{att}}}$. The attention mechanism can be enhanced with multi-head structure in which $h_\mathrm{att}$ linear transformation layers are used to project each input representation to different feature spaces. The output representations of equation \ref{eq5} from all heads are then concatenated as the final output.
We first enable models to process all slot representations together rather than separately as in previous DST models \cite{ramadan2018large, wu-etal-2019-transferable}. This strategy allows our models to explicitly learn dependencies between all pairs of slots. 
Many pairs of slots could exhibit correlation such as time-wise relation (``departure\_time" and ``arrival\_time"). We obtain $Z^\mathrm{dst}_{SS} = \mathrm{Att}(Z_S, Z_S) \in \mathbb{R}^{\|S\| \times d}$.

\noindent We incorporate the dialogue information by learning dependencies between each slot representation and each token in the dialogue history. Previous approaches such as \cite{budzianowski-vulic-2019-hello} consider all dialogue history as a single sequence but we separate them into two inputs because the information in $X_\mathrm{utt}$ is usually more important to generate responses while $X_\mathrm{ctx}$ includes more background information.
We then obtain $Z^\mathrm{dst}_{S,\mathrm{ctx}} = \mathrm{Att}(Z_\mathrm{ctx}, Z^\mathrm{dst}_{SS}) \in \mathbb{R}^{\|S\| \times d}$ and $Z^\mathrm{dst}_{S,\mathrm{utt}} = \mathrm{Att}(Z_\mathrm{utt}, Z^\mathrm{dst}_\mathrm{S,ctx}) \in \mathbb{R}^{\|S\| \times d}$. 

\noindent Following \cite{lei2018sequicity}, we  incorporate dialogue state of the previous turn $B_{t-1}$ which is a more compact representation of dialogue context. Hence, we can replace the full dialogue context to only $R_{t-1}$ as the remaining part is represented in $B_{t-1}$. 
This approach avoids taking in all dialogue history and is scalable as the conversation grows longer.  
We add the attention layer to obtain $Z^\mathrm{dst}_{S,\mathrm{st}} = \mathrm{Att}(Z^\mathrm{prev}_\mathrm{st}, Z^\mathrm{dst}_{S,\mathrm{ctx}}) \in \mathbb{R}^{\|S\| \times d}$ (See Figure \ref{fig:model}). 
%Using dialogue state of the previous turn provides a more information-intensive input yet requires less memory than processing a full-length dialogue context input.
We further improve the feature representations by repeating the attention sequence over $N^\mathrm{dst}_S$ times. 
%At the end of each round, the representation $Z^\mathrm{dst}_{S,\mathrm{utt}}$ is used as $Z_2$ in equations \ref{eq1} to \ref{eq5} in the next attention block. 
We denote the final output $Z^\mathrm{dst}_S$.

\noindent \textbf{Domain-level DST}.
We adopt a similar architecture to learn domain-level representations. 
The representations learned in this module exhibit global information while slot-level representations contain local dependencies to decode multi-domain dialogue states.  
First, we enable the domain-level DST to capture dependencies between all pairs of domains. 
For example, some domains such as ``taxi'' are typically paired with other domains such as ``attraction'', but usually not with the ``train'' domain. 
We then obtain $Z^\mathrm{dst}_{DD} = \mathrm{Att}(Z_{D}, Z_{D}) \in \mathbb{R}^{\|D\| \times d}$.

\noindent We then allow models to capture dependencies between each domain representation and each token in dialogue context and current user utterance.
 By segregating dialogue context and current utterance, our models can potentially detect changes of dialogue domains from past turns to the current turn. 
 Especially in multi-domain dialogues, users can switch from one domain to another and the next system response should address the latest domain.
We then obtain $Z^\mathrm{dst}_{D,\mathrm{ctx}} = \mathrm{Att}(Z_\mathrm{ctx}, Z^\mathrm{dst}_{DD}) \in \mathbb{R}^{\|D\| \times d}$ and $Z^\mathrm{dst}_{D,\mathrm{utt}} = \mathrm{Att}(Z_\mathrm{utt}, Z^\mathrm{dst}_{D,\mathrm{ctx}}) \in \mathbb{R}^{\|D\| \times d}$ sequentially.
%Different from slot-level DST, we do not use dialogue state of the previous turn as input because we expect global dependencies in domain representations are easier to detect. 
Similar to the slot-level module, we refine feature representations over $N^\mathrm{dst}_D$ times and denote the final output as $Z^\mathrm{dst}_D$.

\noindent \textbf{Domain-Slot DST}.
We combined domain and slot representations by expanding the tensors to identical dimensions i.e. $\|D\| \times \|S\| \times d$.
We then apply Hadamard product, resulting in domain-slot joint features $Z^\mathrm{dst}_{DS} \in \mathbb{R}^{\|D\| \times \|S\| \times d}$. 
We then apply a self-attention layer to allow learning of dependencies between joint domain-slot features: $Z^\mathrm{dst} = \mathrm{Att}(Z^\mathrm{dst}_{DS}, Z^\mathrm{dst}_{DS}) \in \mathbb{R}^{\|D\| \times \|S\| \times d}$.
 In this attention, we mask the intermediate representations in positions of invalid domain-slot pairs. 
 %For example, there is no \textit{departure} slot in the \textit{hotel} domain and its position is masked in each $z^{out}_{DS}$. 
 Compared to previous work such as \cite{wu-etal-2019-transferable}, we adopt a \textit{late fusion} method whereby domain and slot representations are integrated in deeper layers.
 \subsubsection{State Generator}
 The representations $Z^\mathrm{dst}$ are used as context-aware representations to decode individual dialogue states. 
 Given a domain index $i$ and slot index $j$, the feature vector $Z_{dst}[i,j,:] \in \mathbb{R}^d$ is used to generate value of the corresponding (domain, slot) pair. 
 The vector is used as an initial hidden state for an RNN decoder to decode an \textit{inform} slot value.
 Given the  $k$-th (domain, slot) pair and decoding step $l$, the output hidden state in each recurrent step $h_{kl}$ is passed through a linear transformation with softmax to obtain output distribution over vocabulary set $V$:
$P^\mathrm{inf}_\mathrm{kl} = \mathrm{Softmax}(h_{kl} W_\mathrm{inf}) \in \mathbb{R}^{\|V\|}$
where $W^\mathrm{inf}_\mathrm{dst} \in \mathbb{R}^{d_\mathrm{rnn} \times \|V\|}$.
For \textit{request} slot of $k$-th (domain,slot) pair, we pass the corresponding vector $Z_{dst}$ vector through a linear layer with sigmoid activation to predict a value of 0 or 1. 
$P^\mathrm{req}_\mathrm{k} = \mathrm{Sigmoid}(Z^{dst}_k W_\mathrm{req})$.
%where $W_\mathrm{req} \in \mathbb{R}^{d \times 1}$.

\noindent \textbf{Optimization}.
The DST is optimized by the cross-entropy loss functions of \textit{inform} and \textit{request} slots: 
\begin{align*}
    \mathcal{L_\mathrm{dst}} = \mathcal{L}_\mathrm{inf} + \mathcal{L}_\mathrm{req} 
    = \sum_{k=1}^{\|DS\|} \sum_{l=1}^{\|Y_k\|} -\log(P^\mathrm{inf}_\mathrm{kl}(y_{kl})) \\
    +  \sum_{k=1}^{\|DS\|}
    -y_k \log(P^\mathrm{req}_k) - (1-y_k)(1- \log(P^\mathrm{req}_k))
\end{align*}

\subsection{Joint Dialogue Act and Response Generator (DARG)}

\noindent \textbf{Database Representations}.
%The decoded dialogue states are used to query the database and obtain the number of resulting entities in each domain. 
Following \cite{budzianowski2018multiwoz}, we create a one-hot vector for each domain $d$: $x^{d}_\mathrm{db} \in \{0,1\}^6 $ and $\sum^6_i x^d_{\mathrm{db},i} = 1$. 
Each position of the vector indicates a number or a range of entities. 
The vectors of all domains are concatenated to create a multi-domain vector $X_\mathrm{db} \in \mathbb{R}^{6 \times \|D\|}$. 
We embed this vector as described in Section \ref{subsec:enc}.
%(i.e. with a matrix $E_\mathrm{db} \in \mathbb{R}^{2 \times d}$), resulting in $Z_\mathrm{db} \in \mathbb{R}^{6 \times \|D\| \times d}$. 

\noindent \textbf{Response Generation}. 
We adopt a stacked-attention architecture that sequentially learns dependencies between each token in target responses with each dialogue component representation. 
First, we obtain $Z^\mathrm{gen}_\mathrm{res} = \mathrm{Att}(Z_\mathrm{res}, Z_\mathrm{res}) \in \mathbb{R}^{L_\mathrm{res} \times d}$. 
This attention layer can learn semantics within the target response to construct a more semantically structured sequence. 
We then use attention to capture dependencies in background information contained in dialogue context and user utterance. 
The outputs are 
$Z^\mathrm{gen}_{\mathrm{ctx}} = \mathrm{Att}(Z_\mathrm{ctx}, Z^\mathrm{gen}_{\mathrm{res}}) \in \mathbb{R}^{L_\mathrm{res} \times d}$ 
and $Z^\mathrm{gen}_{\mathrm{utt}} = \mathrm{Att}(Z_\mathrm{utt}, Z^\mathrm{gen}_{\mathrm{ctx}}) \in \mathbb{R}^{L_\mathrm{res} \times d}$ sequentially.

\noindent To incorporate information of dialogue states and DB results, we apply attention steps to capture dependencies between each response token representation and state or DB representation. 
Specifically, we first obtain $Z^\mathrm{gen}_{\mathrm{dst}} = \mathrm{Att}(Z^\mathrm{dst}, Z^\mathrm{gen}_{\mathrm{utt}}) \in \mathbb{R}^{L_\mathrm{res} \times d}$.
In the context-to-text setting, as we directly use the ground-truth dialogue states, we simply replace $Z^\mathrm{dst}$ with ${Z}^\mathrm{curr}_\mathrm{st}$.
Then we obtain $Z^\mathrm{gen}_{\mathrm{db}} = \mathrm{Att}(Z_\mathrm{db}, Z^\mathrm{gen}_{\mathrm{dst}}) \in \mathbb{R}^{L_\mathrm{res} \times d}$.
These attention layers capture the information needed to generate tokens that are towards task completion and supplement the contextual cues obtained in previous attention layers. 
We let the models to progressively capture these dependencies for $N^\mathrm{gen}$ times and denote the final output as $Z^\mathrm{gen}$. 
The final output is passed to a linear layer with softmax activation to decode system responses auto-regressively:
$P^\mathrm{res} = \mathrm{Softmax}(Z^\mathrm{gen} W_\mathrm{gen}) \in \mathbb{R}^{L_\mathrm{res} \times \|V_\mathrm{res}\|}$

\noindent \textbf{Dialogue Act Modeling}.
We couple response generation with dialogue act modeling by learning a latent variable $Z_\mathrm{act} \in \mathbb{R}^{d}$. 
We place the vector in the first position of $Z_\mathrm{res}$, resulting in $Z_\mathrm{res+act} \in \mathbb{R}^{(L_\mathrm{res}+1) \times d}$.
We then pass this tensor to the same stacked attention layers as above. 
By adding the latent variable in the first position, we allow our model to semantically condition all downstream tokens from second position, i.e. all tokens in the target response, on this latent variable. 
The output representation of the latent vector i.e. first row in $Z^\mathrm{gen}$, incorporates contextual signals accumulated from all attention layers and is used to predict dialogue acts. 
We denote this representation as $Z^\mathrm{gen}_\mathrm{act}$ and pass it through a linear layer to obtain a multi-hot encoded tensor. We apply Sigmoid on this tensor to classify each dialogue act as 0 or 1: 
$P^\mathrm{act} = \mathrm{Sigmoid}(Z^\mathrm{gen}_\mathrm{act} W_\mathrm{act}) \in \mathbb{R}^{\|A\|}$.

\begin{table}[]
\small
\centering
\begin{tabular}{llll}
\hline 
\multicolumn{1}{c}{\multirow{2}{*}{\textbf{Domain}}} & \multicolumn{3}{c}{\textbf{\#dialogues}}                                                             \\ \cmidrule(lr){2-4} 
\multicolumn{1}{c}{}                                 & \multicolumn{1}{c}{\textbf{train}} & \multicolumn{1}{c}{\textbf{val}} & \multicolumn{1}{c}{\textbf{test}} \\ \hline
\textbf{Restaurant}                                  & 3,817                              & 438                              & 437                               \\
\textbf{Hotel}                                       & 3,387                              & 416                              & 394                               \\
\textbf{Attraction}                                  & 2,718                              & 401                              & 396                               \\
\textbf{Train}                                       & 3,117                              & 484                              & 495                               \\
\textbf{Taxi}                                        & 1,655                              & 207                              & 195                               \\
\textbf{Police}                                      & 245                                & 0                                & 0                                 \\
\textbf{Hospital}                                    & 287                                & 0                                & 0  \\ \hline                              
\end{tabular}
\caption{Summary of MultiWOZ dataset \citep{budzianowski2018multiwoz} by domain}
\label{tab:domain_data}
\end{table}

\noindent \textbf{Optimization}.
The response generator is jointly trained by the cross-entropy loss functions of generated responses and dialogue acts: 
\begin{align*}
    \mathcal{L}_\mathrm{gen}=\mathcal{L}_\mathrm{res} + \mathcal{L}_\mathrm{act}
    = \sum_{l=1}^{\|Y_\mathrm{res}\|} -\log(P^\mathrm{res}_l(y_l)) \\
    + \sum_{a=1}^{\|A\|} -y_a \log(P^\mathrm{act}_a) - (1-y_a)(1- \log(P^\mathrm{act}_a))
\end{align*}

\section{Experiments}
\begin{table*}[htbp]
    \begin{minipage}{.4\linewidth}
      \centering
      \small
        \begin{tabular}{ll}
        \hline
        \textbf{Model}          & \textbf{Joint Acc.}           \\
        \hline
        HJST \citep{eric2019multiwoz}          & 35.55\%            \\
        DST Reader \citep{gao2019dialog}    & 36.40\%            \\
        TSCP \citep{lei2018sequicity}         & 37.12\%            \\
        FJST \citep{eric2019multiwoz}          & 38.00\%            \\
        HyST  \citep{goel2019hyst}         & 38.10\%            \\
        TRADE \citep{wu-etal-2019-transferable}         & 45.60\%            \\
        NADST \citep{Le2020Non-Autoregressive} &  49.04\%   \\
        DSTQA \citep{zhou2019multi} & 51.17\% \\
        SOM-DST \citep{kim-etal-2020-efficient} &  \textbf{53.01\%}   \\
        \textbf{BDST (Ours)}  & 49.55\% \\
        \hline
        \end{tabular}
        \caption{Evaluation of DST on MultiWOZ2.1}
        \label{tab:dst}
    \end{minipage}%
    \begin{minipage}{.6\linewidth}
      \centering
      \small
        \begin{tabular}{llll}
        \hline
        \textbf{Model}    & \textbf{Inform}  & \textbf{Success} & \textbf{BLEU}  \\\hline
        \vspace{0.035in}
        Baseline \citet{budzianowski2018multiwoz}          & 71.29\%          & 60.96\%          & 18.80          \\\vspace{0.035in}
        TokenMoE \citep{pei2019modular}         & 75.30\%          & 59.70\%          & 16.81          \\\vspace{0.035in}
        HDSA \citep{chen2019semantically}             & 82.90\%          & 68.90\%          & \textbf{23.60} \\\vspace{0.035in}
        Structured Fusion \citep{mehri2019structured} & 82.70\%          & 72.10\%          & 16.34          \\\vspace{0.035in}
        LaRL \citep{zhao-etal-2019-rethinking}           & 82.78\%          & \textbf{79.20\%} & 12.80          \\\vspace{0.035in}
        GPT2 \citep{budzianowski-vulic-2019-hello}           & 70.96\%          & 61.36\% & 19.05          \\\vspace{0.035in}
        DAMD \citep{zhang2019task} & \textbf{89.50\%} & 75.80\% & 18.30 \\\vspace{0.035in}
        \textbf{DARG (Ours)}     & 87.80\% & 73.60\%          & 18.80         \\ \hline
        \end{tabular}
        \caption{Evaluation of context-to-text task on MultiWOZ2.0.}
        \label{tab:c2t}
    \end{minipage} 
    %\vspace{-0.1in}
\end{table*}

\subsection{Dataset}
We evaluate our models with the multi-domain dialogue corpus MultiWOZ 2.0 \citep{budzianowski2018multiwoz} and 2.1 \citep{eric2019multiwoz} (The latter includes corrected state labels for the DST task).
From the dialogue state annotation of the training data, we identified all possible domains and slots. We identified $\|D\| = 7$ domains and $\|S\| = 30$ slots, including 19 \textit{inform} slots and 11 \textit{request} slots. 
We also identified $\|A\|=32$ acts. 
The corpus includes 8,438 dialogues in the training set and 1,000 in each validation and test set.
We present a summary of the dataset in Table \ref{tab:domain_data}.
For additional information of data pre-processing procedures, domains, slots, and entity DBs, please refer to Appendix \ref{app:data}. 

\subsection{Experiment Setup}
We select $d=256$, $h_\mathrm{att}=8$,  $N^\mathrm{dst}_S=N^\mathrm{dst}_D=N^\mathrm{gen}=3$.
We employed dropout \citep{srivastava2014dropout} of $0.3$ and label smoothing \citep{szegedy2016rethinking} on target system responses during training.
%at all network layers except the linear layers in the generative components. 
We adopt a teacher-forcing training strategy by simply using the ground-truth inputs of dialogue state of the previous turn and the gold DB representations. 
During inference in DST and end-to-end tasks, we decode system responses sequentially turn by turn, using the previously decoded state as input in the current turn, and at each turn, using the new predicted state to query DBs. 
%For the context-to-text generation task, ground-truth dialogue states and DBs are used during both training and inference. 
We train all networks with Adam optimizer \citep{kingma2014adam} and a decaying learning rate schedule.
%We used batch size $32$ and tuned the \textit{warmup\_steps} from 10K to 15K training steps.
All models are trained up to $30$ epochs and the best models are selected based on validation loss. 
We used a greedy approach to decode all slots and a beam search with beam size $5$.
%and a length penalty of $1.0$ to decode responses. 
%The maximum length is set to 10 tokens for each slot and $20$ for system responses. %Our models are implemented using PyTorch \citep{paszke2017automatic}.
To evaluate the models, we use the following metrics: Joint Accuracy and Slot Accuracy \citep{henderson2014word}, Inform and Success \citep{wen2016network}, and BLEU score \citep{papineni2002bleu}.
As suggested by \citet{liu-etal-2016-evaluate}, human evaluation, even though popular in dialogue research, might not be necessary in tasks with domain constraints such as MultiWOZ. 
We implemented all models using Pytorch and will release our code on github\footnote{\url{https://github.com/henryhungle/UniConv}}.

\subsection{Results}
%\vspace{-0.2cm}
\begin{table*}[ht!]
\centering
\small
\begin{tabular}{llllll}
\hline
\textbf{Model} & \textbf{Joint Acc} & \textbf{Slot Acc} & \textbf{Inform}  & \textbf{Success} & \textbf{BLEU}  \\\hline
TSCP (L=8) \cite{lei2018sequicity}      & 31.64\%            & 95.53\%           & 45.31\%          & 38.12\%          & 11.63          \\
TSCP (L=20) \cite{lei2018sequicity} & 37.53\%            & 96.23\%           & 66.41\%          & 45.32\%          & 15.54          \\
HRED-TS \cite{peng2019teacher} & -            & -          & 70.00\%          & 58.00\%          & 17.50          \\
Structured Fusion \citep{mehri2019structured} & -            & -          & 73.80\%          & 58.60\%          & 16.90          \\

DAMD \cite{zhang2019task} & -            & -          & \textbf{76.30\%}          & 60.40\%          & 16.60          \\
\textbf{UniConv (Ours)}  & \textbf{50.14\%}   & \textbf{97.30\%}  & 72.60\% & \textbf{62.90\%} & \textbf{19.80} \\\hline
\end{tabular}
%}
\caption{Evaluation on MultiWOZ2.1 in the end-to-end setting.}
\label{tab:e2e}
%\vspace{-0.1in}
\end{table*}

%We report our best results in each task and ablation analysis with different model variants. 

\noindent \textbf{DST}.
We test our state tracker (i.e. using only $\mathcal{L}_\mathrm{dst}$) and compare the performance with the baseline models in Table \ref{tab:dst} (Refer to Appendix \ref{app:baselines} 
for description of DST baselines). 
%Our model achieves the SOTA performance in MultiWOZ2.1 corpus. 
%By leveraging on dialogue context signals through independent attention modules at domain level and slot level, our DST can generate slot values more accurately. 
Our model can outperform fixed-vocabulary approaches such as HJST and FJST, showing the advantage of generating unique slot values rather than relying on a slot ontology with a fixed set of candidates. 
DST Reader model \cite{gao2019dialog} does not perform well and we note that many slot values are not easily expressed as a text span in source text inputs.
DST approaches that separate domain and slot representations such as TRADE \cite{wu-etal-2019-transferable} reveal competitive performance. However, our approach has better performance as we adopt a \textit{late fusion} strategy to explicitly obtain more fine-grained contextual dependencies in each domain and slot representation. 
In this aspect, our model is related to TSCP \cite{lei2018sequicity} which decodes output state sequence auto-regressively. However, TSCP attempts to learn domain and slot dependencies implicitly and the model is limited by selecting the maximum output state length (which can vary significantly in multi-domain dialogues). 
%We also noted that DST performance improves when our models are trained as an end-to-end system (See Table \ref{tab:e2e}). This can be explained as additional supervision from system responses not only contributes to learn a good response generation network but also positively impact the DST network. 
%Additional DST results of individual domains can be seen in the Supplementary Material.

\noindent \textbf{Context-to-Text Generation}.
We compare with existing baselines in Table \ref{tab:c2t} (Refer to Appendix \ref{app:baselines} 
for description of the baseline models). 
Our model achieves very competitive Inform, Success, and BLEU scores.
%By modelling the dialogue acts as a prior which is then used to semantically condition each token in the target response, our models are able to explicitly leverage the dialogue act signals in addition to contextual cues from input components. 
Compared to TokenMOE \cite{pei2019modular}, our single model can outperform multiple domain-specific dialogue agents as each attention module can sufficiently learn contextual features of multiple domains. 
Compared to HDSA \cite{chen2019semantically} which uses a graph structure to represent \textit{acts}, our approach is simpler yet able to outperform HDSA in Inform score. 
Our work is related to Structured Fusion \cite{mehri2019structured} as we incorporate intermediate representations during decoding. However, our approach does not rely on pre-training individual sub-modules but simultaneously learning both act representations and predicting output tokens.
Similarly, our stacked attention architecture can achieve good performance in BLEU score, competitively with a GPT-2 based model \cite{budzianowski-vulic-2019-hello}, while consistently improve other metrics. 
For completion, we tested our models on MultiWOZ2.1 and achieved similar results: 87.90\% \textit{Inform}, 72.70\% \textit{Success}, and 18.52 BLEU score. 
Future work may further improve \textit{Success} by optimizing the models towards a higher success rate using strategies such as LaRL \cite{zhao-etal-2019-rethinking}. 
Another direction is a data augmentation approach such as DAMD \cite{zhang2019task} which achieves significant performance gain in this task.
\begin{table*}[ht!]
\small
\resizebox{1.0\textwidth}{!} {
\begin{tabular}{ccccccccclllll}
\hline
\# & \textbf{$X_\mathrm{ctx}$} & \textbf{$B_{t-1}$} & \textbf{$N^
\mathrm{dst}_{S}$} & \textbf{$N^\mathrm{dst}_\mathrm{D}$} & \textbf{$N^\mathrm{gen}$} & \textbf{$\mathcal{L}_\mathrm{act}$} & \textbf{$d$} & \textbf{$h_\mathrm{att}$} & \multicolumn{1}{c}{\textbf{Joint Acc.}} & \multicolumn{1}{c}{\textbf{Slot Acc.}} & \multicolumn{1}{c}{\textbf{Inform}} & \multicolumn{1}{c}{\textbf{Success}} & \multicolumn{1}{c}{\textbf{BLEU}} \\
\hline 
A1 & $R_{t-1}$                    & \checkmark               & 3                                  & 3                                  & 0               &                 & 256        & 8               & \textbf{49.55\%}                        & 97.32\%                                & -                                   & -                                    & -                                 \\
A2& $R_{t-1}$                    & \checkmark               & 3                                  & 0                                  & 0               &                 & 256        & 8               & 47.91\%                                 & 97.25\%                                & -                                   & -                                    & -                                 \\
A3& $R_{t-1}$                    & \checkmark               & 2                                  & 2                                  & 0               &                 & 256        & 8               & 47.80\%                                 & 97.22\%                                & -                                   & -                                    & -                                 \\
A4& $R_{t-1}$                    & \checkmark               & 1                                  & 1                                  & 0               &                 & 256        & 8               & 46.20\%                                 & 97.08\%                                & -                                   & -                                    & -                                 \\
A5&$(U,R)_{1:t-1}$                  & \checkmark               & 3                                  & 3                                  & 0               &                 & 256        & 8               & 49.20\%                                 & \textbf{97.34\%}                       & -                                   & -                                    & -                                 \\
%All                  &                 & 3                                  & 3                                  & 0               &                 & 256        & 8               & 36.38\%                                 & 92.52\%                                & -                                   & -                                    & -                                 \\
\hline
B1 & $R_{t-1}$                    &                 & 0                                  & 0                                  & 3               & \checkmark               & 256        & 8               & -                                       & -                                      & \textbf{87.90\%}                    & \textbf{72.70\%}                     & 18.52                             \\
B2 & $R_{t-1}$                    &                 & 0                                  & 0                                  & 3               &                 & 256        & 8               & -                                       & -                                      & 82.70\%                             & 70.60\%                              & 18.51                             \\
B3 & $(U,R)_{1:t-1}$                   &                 & 0                                  & 0                                  & 3               & \checkmark               & 256        & 8               & -                                       & -                                      & 87.14\%                             & 71.52\%                              & \textbf{18.90}                    \\
B4 & $R_{t-1}$                    &                 & 0                                  & 0                                  & 2               & \checkmark               & 256        & 8               & -                                       & -                                      & 81.60\%                             & 66.40\%                              & 18.48                             \\
B5 & $R_{t-1}$                    &                 & 0                                  & 0                                  & 1               & \checkmark               & 256        & 8               & -                                       & -                                      & 77.70\%                             & 62.80\%                              & 18.50                             \\
\hline
C1 & $R_{t-1}$                    & \checkmark               & 3                                  & 3                                  & 3               & \checkmark               & 256        & 8               & \textbf{50.14\%}                                 & \textbf{97.30}\%                                & \textbf{72.60\%}                             & \textbf{62.90\%}                              & 19.80                             \\
C2 & $R_{t-1}$                    & \checkmark               & 3                                  & 3                                  & 3               & \checkmark               & 128        & 8               &     45.70\%                                    &     97.00\%                                   &   67.40\%                                  &   58.30\%                                   &         \textbf{19.90}                          \\
C3 & $R_{t-1}$                    & \checkmark               & 3                                  & 3                                  & 3               & \checkmark               & 256        & 4               &      47.30\%                                   &      97.10\%                                  &        68.70\%                             &       57.10\%                               &        19.60                           \\
C4 & $R_{t-1}$                    & \checkmark               & 3                                  & 3                                  & 3               & \checkmark               & 256        & 2               &        45.90\%                                 &       97.00\%                                 &        66.10\%                             &    55.60\%                                  &        19.80                           \\
C5 & $R_{t-1}$                    & \checkmark               & 3                                  & 3                                  & 3               & \checkmark               & 256        & 1               &              43.30\%                           &    96.70\%                                    &      62.30\%                               &      52.60\%                                &  \textbf{19.90}    \\        \hline                    
\end{tabular}
}
\caption{Ablation analysis on the MultiWOZ2.1 in DST (top), context-to-text (middle), and end-to-end (bottom).}
\label{tab:ablation}
%\vspace{-0.2in}
\end{table*}

\noindent \textbf{End-to-End}.
%In this setting, we compare our model performance on the joint task of DST and context-to-text generation (See the Supplementary Material for a description of baseline models). 
From Table \ref{tab:e2e}, our model outperforms existing baselines in all metrics except the Inform score (See Appendix \ref{app:baselines} 
for a description of baseline models).
In TSCP \cite{lei2018sequicity}, increasing the maximum dialogue state span $L$ from 8 to 20 tokens helps to improve the DST performance, but also increases the training time significantly. 
Compared with HRED-TS \cite{peng2019teacher}, our single model generates better responses in all domains without relying on multiple domain-specific teacher models. 
We also noted that the performance of DST improves in contrast to the previous DST task. This can be explained as additional supervision from system responses not only contributes to learn a natural response but also positively impact the DST component. 
%We experimented with other baselines along the line of research of sequence-to-sequence framework 
Other baseline models such as \cite{eric2017copy, wu2019global} present challenges in the MultiWOZ benchmark as the models could not fully optimize due to the large scale entity memory.
For example, following GLMP \cite{wu2019global}, the \textit{restaurant} domain alone has over 1,000 memory tuples of \textit{(Subject, Relation, Object)}.
%Our dialogue systems, including the Multi-Level State Tracker and Stacked Attention Response Generator, achieve better performance in all evaluation measures. 

\noindent \textbf{Ablation}. 
We conduct a comprehensive ablation analysis with several model variants in Table \ref{tab:ablation} and have the following observations: 
\begin{itemize}
\item The model variant with a single-level DST (by considering $S=DS$ and $N^\mathrm{dst}_D=0$) (Row A2) performs worse than the Bi-level DST (Row A1). 
In addition, using the dual architecture also improves the latency in each attention layers as typically $\|D\| + \|S\| \ll \|DS\|$.
The performance gap also indicates the potential of separating global and local dialogue state dependencies by domain and slot level. 
%\vspace{-0.05in}
\item Using $B_{t-1}$ and only the last user utterance as the dialogue context (Row A1 and B1) performs as well as using $B_{t-1}$ and a full-length dialogue history (Row A5 and B3).
This demonstrates that the information from the last dialogue state is sufficient to represent the dialogue history up to the last user utterance. 
One benefit from not using the full dialogue history is that it reduces the memory cost as the number of tokens in a full-length dialogue history is much larger than that of a dialogue state (particularly as the conversation evolves over many turns).
%\vspace{-0.05in}
\item We note that removing the loss function to learn the dialogue act latent variable (Row B2) can hurt the generation performance, especially by the task completion metrics \emph{Inform} and \emph{Success}. 
This is interesting as we expect dialogue acts affect the general semantics of output sentences, indicated by BLEU score, rather than the model ability to retrieve correct entities. 
This reveals the benefit of our approach.
By enforcing a semantic condition on each token of the target response, the model can facility the dialogue flow towards successful task completion. 
%\vspace{-0.05in}
\item In both state tracker and response generator modules, we note that learning feature representations through deeper attention networks can improve the quality of predicted states and system responses.
This is consistent with our DST performance as compared to baseline models of shallow networks. 
%\vspace{-0.05in}
\item Lastly, in the end-to-end task, our model achieves better performance as the number of attention heads increases, by learning more high-resolution dependencies. 
%We note that however, the BLEU metric changes only slightly with different parameter setting, suggesting a potential direction to improve our models.
\end{itemize}
%\vspace{-0.05in}

\section{Domain-dependent Results}
\label{app:domain_dependent_results}
\noindent \textbf{DST}. 
For state tracking, the metrics are calculated for domain-specific slots of the corresponding domain at each dialogue turn. 
We also report the DST separately for multi-domain and single-domain dialogues to evaluate the challenges in multi-domain dialogues and our DST performance gap as compared to single-domain dialogues. 
From Table \ref{tab:add_result_dst}, our DST performs consistently well in the 3 domains \textit{attraction, restaurant}, and \textit{train} domains.
However, the performance drops in the \textit{taxi} and \textit{hotel} domain, significantly in the \textit{taxi} domain. 
We note that dialogues with the \textit{taxi} domain is usually not single-domain but typically entangled with other domains. 
%For example, the users tend to converse about a hotel or an attraction in the beginning and only request a taxi towards the end of a dialogue.
%This increases the complexity of dialogue states in the \textit{taxi} domain.
%as the dialogue history, in this case, extends over many turns and contains many more references to past user utterances. 
%We will investigate further to identify and address these challenges in future work.
Secondly, we observe that there is a significant performance gap of about 10 points absolute score between DST performances in single-domain and multi-domain dialogues. 
State tracking in multi-domain dialogues is, hence, could be further improved to boost the overall performance. 

\begin{table}[htbp]
\centering
\small
\begin{tabular}{lll}
\hline
\textbf{Domain} & \textbf{Joint Acc} & \textbf{Slot Acc} \\\hline
Multi-domain      & 48.40\%            & 97.14\%           \\
Single-domain      & 59.63\%            & 98.36\%           \\
\hline
Attraction      & 66.76\%            & 98.94\%           \\
Hotel           & 47.86\%            & 97.54\%           \\
Restaurant      & 65.11\%            & 98.68\%           \\
Taxi            & 30.84\%            & 96.86\%           \\
Train           & 63.77\%            & 98.53\%      \\\hline    
\end{tabular}
\caption{DST results on MultiWOZ2.1 by domains.
%In each domain, the metrics are calculated for domain-specific slots of the corresponding domain at each dialogue turn.
}
\label{tab:add_result_dst}
%\vspace{-0.4cm}
\end{table}

\noindent \textbf{Context-to-Text Generation}
For this task, we calculated the metrics for single-domain dialogues of the corresponding domain (as Inform and Success are computed per dialogue rather than per turn). 
We do not report the \textit{Inform} metric of the \textit{taxi} domain because no DB was available for this domain.
From Table \ref{tab:add_result_c2t}, we observe some performance gap between \textit{Inform} and \textit{Success} scores on multi-domain dialogues and single-domain dialogues.
However, in terms of BLEU score, our model performs better with multi-domain dialogues. 
This could be caused by the data bias in MultiWOZ corpus as the majority of dialogues in this corpus is multi-domain. Hence, our models capture the semantics of multi-domain dialogue responses better than single-domain responses. 
For domain-specific results, we note that our models perform not as well as other domains in \textit{attraction} and \textit{taxi} domains in terms of \textit{Success} score. 
%The model also does not perform well in BLEU score in the \textit{taxi} domain, suggesting challenges remained in dialogues of this particular domain.

\begin{table}[htbp]
\centering
\small
\begin{tabular}{llll}
\hline
\textbf{Domain} & \textbf{Inform} & \textbf{Success} & \textbf{BLEU} \\\hline
Multi-domain      & 85.01\%            & 68.86\%           &18.68\\
Single-domain      & 97.79\%            & 85.84\%          &17.62\\
\hline
Attraction      & 91.67\%            & 66.67\%          & 19.17 \\
Hotel           & 97.01\%            & 91.04\%          & 16.55\\
Restaurant      & 96.77\%            & 88.71\%         &  19.88 \\
Taxi            & -            & 78.85\%         &  13.85 \\
Train           & 99.10\%            & 87.88\%      & 18.14 \\\hline    
\end{tabular}
\caption{Context-to-text generation results on MultiWOZ2.1. by domains.}
\label{tab:add_result_c2t}
%\vspace{-0.4cm}
\end{table}

\noindent Additionally, we report qualitative analysis and the insights can be seen in Appendix \ref{app:qual_res}.

%\vspace{-0.2cm}
\section{Conclusion} 
%\vspace{-0.2cm}
We proposed \emph{UniConv}, a novel unified neural architecture of conversational agents for Multi-domain Task-oriented Dialogues, which jointly trains (1) a Bi-level State Tracker to capture dependencies in both domain and slot levels simultaneously, and (2) a Joint Dialogue Act and Response Generator to model dialogue act latent variable and semantically conditions output responses with contextual cues. The promising performance of UniConv on the MultiWOZ benchmark (including three tasks: DST, context-to-text generation, and end-to-end dialogues) validates the efficacy of our method. 

\section*{Acknowledgments}

We thank all reviewers for their insightful feedback to the manuscript of this paper. 
The first author of this paper is supported by the Agency for Science, Technology and Research (A*STAR) Computing and Information Science scholarship.

\bibliography{anthology,emnlp2020}

\begin{thebibliography}{53}
\expandafter\ifx\csname natexlab\endcsname\relax\def\natexlab#1{#1}\fi

\bibitem[{Ba et~al.(2016)Ba, Kiros, and Hinton}]{ba2016layer}
Jimmy~Lei Ba, Jamie~Ryan Kiros, and Geoffrey~E Hinton. 2016.
\newblock Layer normalization.
\newblock \emph{arXiv preprint arXiv:1607.06450}.

\bibitem[{Bordes et~al.(2016)Bordes, Boureau, and Weston}]{bordes2016learning}
Antoine Bordes, Y-Lan Boureau, and Jason Weston. 2016.
\newblock Learning end-to-end goal-oriented dialog.
\newblock \emph{arXiv preprint arXiv:1605.07683}.

\bibitem[{Budzianowski and Vuli{\'c}(2019)}]{budzianowski-vulic-2019-hello}
Pawe{\l} Budzianowski and Ivan Vuli{\'c}. 2019.
\newblock \href {https://doi.org/10.18653/v1/D19-5602} {Hello, it{'}s {GPT}-2 -
  how can {I} help you? towards the use of pretrained language models for
  task-oriented dialogue systems}.
\newblock In \emph{Proceedings of the 3rd Workshop on Neural Generation and
  Translation}, pages 15--22, Hong Kong. Association for Computational
  Linguistics.

\bibitem[{Budzianowski et~al.(2018)Budzianowski, Wen, Tseng, Casanueva, Stefan,
  Osman, and Ga{\v{s}}i\'c}]{budzianowski2018multiwoz}
Pawe{\l} Budzianowski, Tsung-Hsien Wen, Bo-Hsiang Tseng, I{\~n}igo Casanueva,
  Ultes Stefan, Ramadan Osman, and Milica Ga{\v{s}}i\'c. 2018.
\newblock Multiwoz - a large-scale multi-domain wizard-of-oz dataset for
  task-oriented dialogue modelling.
\newblock In \emph{Proceedings of the 2018 Conference on Empirical Methods in
  Natural Language Processing (EMNLP)}.

\bibitem[{Chen et~al.(2019)Chen, Chen, Qin, Yan, and
  Wang}]{chen2019semantically}
Wenhu Chen, Jianshu Chen, Pengda Qin, Xifeng Yan, and William~Yang Wang. 2019.
\newblock \href {https://doi.org/10.18653/v1/P19-1360} {Semantically
  conditioned dialog response generation via hierarchical disentangled
  self-attention}.
\newblock In \emph{Proceedings of the 57th Annual Meeting of the Association
  for Computational Linguistics}, pages 3696--3709, Florence, Italy.
  Association for Computational Linguistics.

\bibitem[{Eric et~al.(2019)Eric, Goel, Paul, Sethi, Agarwal, Gao, and
  Hakkani-Tur}]{eric2019multiwoz}
Mihail Eric, Rahul Goel, Shachi Paul, Abhishek Sethi, Sanchit Agarwal, Shuyag
  Gao, and Dilek Hakkani-Tur. 2019.
\newblock Multiwoz 2.1: Multi-domain dialogue state corrections and state
  tracking baselines.
\newblock \emph{arXiv preprint arXiv:1907.01669}.

\bibitem[{Eric et~al.(2017)Eric, Krishnan, Charette, and Manning}]{eric2017key}
Mihail Eric, Lakshmi Krishnan, Francois Charette, and Christopher~D. Manning.
  2017.
\newblock \href {https://doi.org/10.18653/v1/W17-5506} {Key-value retrieval
  networks for task-oriented dialogue}.
\newblock In \emph{Proceedings of the 18th Annual {SIG}dial Meeting on
  Discourse and Dialogue}, pages 37--49, Saarbr{\"u}cken, Germany. Association
  for Computational Linguistics.

\bibitem[{Eric and Manning(2017)}]{eric2017copy}
Mihail Eric and Christopher Manning. 2017.
\newblock \href {https://www.aclweb.org/anthology/E17-2075} {A copy-augmented
  sequence-to-sequence architecture gives good performance on task-oriented
  dialogue}.
\newblock In \emph{Proceedings of the 15th Conference of the {E}uropean Chapter
  of the Association for Computational Linguistics: Volume 2, Short Papers},
  pages 468--473, Valencia, Spain. Association for Computational Linguistics.

\bibitem[{Gangi~Reddy et~al.(2019)Gangi~Reddy, Contractor, Raghu, and
  Joshi}]{gangi-reddy-etal-2019-multi}
Revanth Gangi~Reddy, Danish Contractor, Dinesh Raghu, and Sachindra Joshi.
  2019.
\newblock \href {https://doi.org/10.18653/v1/N19-1375} {Multi-level memory for
  task oriented dialogs}.
\newblock In \emph{Proceedings of the 2019 Conference of the North {A}merican
  Chapter of the Association for Computational Linguistics: Human Language
  Technologies, Volume 1 (Long and Short Papers)}, pages 3744--3754,
  Minneapolis, Minnesota. Association for Computational Linguistics.

\bibitem[{Gao et~al.(2019)Gao, Sethi, Agarwal, Chung, and
  Hakkani-Tur}]{gao2019dialog}
Shuyang Gao, Abhishek Sethi, Sanchit Agarwal, Tagyoung Chung, and Dilek
  Hakkani-Tur. 2019.
\newblock \href {https://doi.org/10.18653/v1/W19-5932} {Dialog state tracking:
  A neural reading comprehension approach}.
\newblock In \emph{Proceedings of the 20th Annual SIGdial Meeting on Discourse
  and Dialogue}, pages 264--273, Stockholm, Sweden. Association for
  Computational Linguistics.

\bibitem[{Goel et~al.(2019)Goel, Paul, and Hakkani-Tür}]{goel2019hyst}
Rahul Goel, Shachi Paul, and Dilek Hakkani-Tür. 2019.
\newblock \href {https://doi.org/10.21437/Interspeech.2019-1863} {{HyST: A
  Hybrid Approach for Flexible and Accurate Dialogue State Tracking}}.
\newblock In \emph{Proc. Interspeech 2019}, pages 1458--1462.

\bibitem[{Henderson et~al.(2014{\natexlab{a}})Henderson, Thomson, and
  Williams}]{henderson2014second}
Matthew Henderson, Blaise Thomson, and Jason~D Williams. 2014{\natexlab{a}}.
\newblock The second dialog state tracking challenge.
\newblock In \emph{Proceedings of the 15th Annual Meeting of the Special
  Interest Group on Discourse and Dialogue (SIGDIAL)}, pages 263--272.

\bibitem[{Henderson et~al.(2014{\natexlab{b}})Henderson, Thomson, and
  Young}]{henderson2014word}
Matthew Henderson, Blaise Thomson, and Steve Young. 2014{\natexlab{b}}.
\newblock Word-based dialog state tracking with recurrent neural networks.
\newblock In \emph{Proceedings of the 15th Annual Meeting of the Special
  Interest Group on Discourse and Dialogue (SIGDIAL)}, pages 292--299.

\bibitem[{Jhunjhunwala et~al.(2020)Jhunjhunwala, Bryant, and
  Shah}]{jhunjhunwala-etal-2020-multi}
Megha Jhunjhunwala, Caleb Bryant, and Pararth Shah. 2020.
\newblock \href {https://www.aclweb.org/anthology/2020.sigdial-1.36}
  {Multi-action dialog policy learning with interactive human teaching}.
\newblock In \emph{Proceedings of the 21th Annual Meeting of the Special
  Interest Group on Discourse and Dialogue}, pages 290--296, 1st virtual
  meeting. Association for Computational Linguistics.

\bibitem[{Kim et~al.(2020)Kim, Yang, Kim, and Lee}]{kim-etal-2020-efficient}
Sungdong Kim, Sohee Yang, Gyuwan Kim, and Sang-Woo Lee. 2020.
\newblock \href {https://doi.org/10.18653/v1/2020.acl-main.53} {Efficient
  dialogue state tracking by selectively overwriting memory}.
\newblock In \emph{Proceedings of the 58th Annual Meeting of the Association
  for Computational Linguistics}, pages 567--582, Online. Association for
  Computational Linguistics.

\bibitem[{Kingma and Ba(2015)}]{kingma2014adam}
Diederick~P Kingma and Jimmy Ba. 2015.
\newblock Adam: A method for stochastic optimization.
\newblock In \emph{International Conference on Learning Representations
  (ICLR)}.

\bibitem[{Le et~al.(2020)Le, Socher, and Hoi}]{Le2020Non-Autoregressive}
Hung Le, Richard Socher, and Steven~C.H. Hoi. 2020.
\newblock \href {https://openreview.net/forum?id=H1e_cC4twS}
  {Non-autoregressive dialog state tracking}.
\newblock In \emph{International Conference on Learning Representations}.

\bibitem[{Lee et~al.(2019)Lee, Lee, and Kim}]{lee2019sumbt}
Hwaran Lee, Jinsik Lee, and Tae-Yoon Kim. 2019.
\newblock \href {https://doi.org/10.18653/v1/P19-1546} {{SUMBT}: Slot-utterance
  matching for universal and scalable belief tracking}.
\newblock In \emph{Proceedings of the 57th Annual Meeting of the Association
  for Computational Linguistics}, pages 5478--5483, Florence, Italy.
  Association for Computational Linguistics.

\bibitem[{Lei et~al.(2018)Lei, Jin, Kan, Ren, He, and Yin}]{lei2018sequicity}
Wenqiang Lei, Xisen Jin, Min-Yen Kan, Zhaochun Ren, Xiangnan He, and Dawei Yin.
  2018.
\newblock Sequicity: Simplifying task-oriented dialogue systems with single
  sequence-to-sequence architectures.
\newblock In \emph{Proceedings of the 56th Annual Meeting of the Association
  for Computational Linguistics (Volume 1: Long Papers)}, pages 1437--1447.

\bibitem[{Li et~al.(2017)Li, Chen, Li, Gao, and Celikyilmaz}]{li2017end}
Xiujun Li, Yun-Nung Chen, Lihong Li, Jianfeng Gao, and Asli Celikyilmaz. 2017.
\newblock \href {https://www.aclweb.org/anthology/I17-1074} {End-to-end
  task-completion neural dialogue systems}.
\newblock In \emph{Proceedings of the Eighth International Joint Conference on
  Natural Language Processing (Volume 1: Long Papers)}, pages 733--743, Taipei,
  Taiwan. Asian Federation of Natural Language Processing.

\bibitem[{Liu and Lane(2017)}]{liu2017end}
Bing Liu and Ian Lane. 2017.
\newblock An end-to-end trainable neural network model with belief tracking for
  task-oriented dialog.
\newblock In \emph{Interspeech 2017}.

\bibitem[{Liu et~al.(2016)Liu, Lowe, Serban, Noseworthy, Charlin, and
  Pineau}]{liu-etal-2016-evaluate}
Chia-Wei Liu, Ryan Lowe, Iulian Serban, Mike Noseworthy, Laurent Charlin, and
  Joelle Pineau. 2016.
\newblock \href {https://doi.org/10.18653/v1/D16-1230} {How {NOT} to evaluate
  your dialogue system: An empirical study of unsupervised evaluation metrics
  for dialogue response generation}.
\newblock In \emph{Proceedings of the 2016 Conference on Empirical Methods in
  Natural Language Processing}, pages 2122--2132, Austin, Texas. Association
  for Computational Linguistics.

\bibitem[{Liu and Perez(2017)}]{liu-perez-2017-gated}
Fei Liu and Julien Perez. 2017.
\newblock \href {https://www.aclweb.org/anthology/E17-1001} {Gated end-to-end
  memory networks}.
\newblock In \emph{Proceedings of the 15th Conference of the {E}uropean Chapter
  of the Association for Computational Linguistics: Volume 1, Long Papers},
  pages 1--10, Valencia, Spain. Association for Computational Linguistics.

\bibitem[{Madotto et~al.(2018)Madotto, Wu, and Fung}]{madotto2018mem2seq}
Andrea Madotto, Chien-Sheng Wu, and Pascale Fung. 2018.
\newblock \href {http://aclweb.org/anthology/P18-1136} {Mem2seq: Effectively
  incorporating knowledge bases into end-to-end task-oriented dialog systems}.
\newblock In \emph{Proceedings of the 56th Annual Meeting of the Association
  for Computational Linguistics (Volume 1: Long Papers)}, pages 1468--1478.
  Association for Computational Linguistics.

\bibitem[{Mehri et~al.(2019)Mehri, Srinivasan, and
  Eskenazi}]{mehri2019structured}
Shikib Mehri, Tejas Srinivasan, and Maxine Eskenazi. 2019.
\newblock \href {https://doi.org/10.18653/v1/W19-5921} {Structured fusion
  networks for dialog}.
\newblock In \emph{Proceedings of the 20th Annual SIGdial Meeting on Discourse
  and Dialogue}, pages 165--177, Stockholm, Sweden. Association for
  Computational Linguistics.

\bibitem[{Papineni et~al.(2002)Papineni, Roukos, Ward, and
  Zhu}]{papineni2002bleu}
Kishore Papineni, Salim Roukos, Todd Ward, and Wei-Jing Zhu. 2002.
\newblock Bleu: a method for automatic evaluation of machine translation.
\newblock In \emph{Proceedings of the 40th annual meeting on association for
  computational linguistics}, pages 311--318. Association for Computational
  Linguistics.

\bibitem[{Pei et~al.(2019)Pei, Ren, and de~Rijke}]{pei2019modular}
Jiahuan Pei, Pengjie Ren, and Maarten de~Rijke. 2019.
\newblock A modular task-oriented dialogue system using a neural
  mixture-of-experts.
\newblock \emph{arXiv preprint arXiv:1907.05346}.

\bibitem[{Peng et~al.(2018)Peng, Li, Gao, Liu, and Wong}]{peng2018deep}
Baolin Peng, Xiujun Li, Jianfeng Gao, Jingjing Liu, and Kam-Fai Wong. 2018.
\newblock \href {https://www.aclweb.org/anthology/P18-1203} {{D}eep {D}yna-{Q}:
  Integrating planning for task-completion dialogue policy learning}.
\newblock In \emph{Proceedings of the 56th Annual Meeting of the Association
  for Computational Linguistics (Volume 1: Long Papers)}, pages 2182--2192,
  Melbourne, Australia. Association for Computational Linguistics.

\bibitem[{Peng et~al.(2017)Peng, Li, Li, Gao, Celikyilmaz, Lee, and
  Wong}]{peng2017composite}
Baolin Peng, Xiujun Li, Lihong Li, Jianfeng Gao, Asli Celikyilmaz, Sungjin Lee,
  and Kam-Fai Wong. 2017.
\newblock \href {https://doi.org/10.18653/v1/D17-1237} {Composite
  task-completion dialogue policy learning via hierarchical deep reinforcement
  learning}.
\newblock In \emph{Proceedings of the 2017 Conference on Empirical Methods in
  Natural Language Processing}, pages 2231--2240, Copenhagen, Denmark.
  Association for Computational Linguistics.

\bibitem[{Peng et~al.(2019)Peng, Huang, Lin, Ji, Chen, and
  Zhang}]{peng2019teacher}
Shuke Peng, Xinjing Huang, Zehao Lin, Feng Ji, Haiqing Chen, and Yin Zhang.
  2019.
\newblock Teacher-student framework enhanced multi-domain dialogue generation.
\newblock \emph{arXiv preprint arXiv:1908.07137}.

\bibitem[{Qin et~al.(2019)Qin, Liu, Che, Wen, Li, and
  Liu}]{qin-etal-2019-entity}
Libo Qin, Yijia Liu, Wanxiang Che, Haoyang Wen, Yangming Li, and Ting Liu.
  2019.
\newblock \href {https://doi.org/10.18653/v1/D19-1013} {Entity-consistent
  end-to-end task-oriented dialogue system with {KB} retriever}.
\newblock In \emph{Proceedings of the 2019 Conference on Empirical Methods in
  Natural Language Processing and the 9th International Joint Conference on
  Natural Language Processing (EMNLP-IJCNLP)}, pages 133--142, Hong Kong,
  China. Association for Computational Linguistics.

\bibitem[{Radford et~al.(2019)Radford, Wu, Child, Luan, Amodei, and
  Sutskever}]{radford2019language}
Alec Radford, Jeff Wu, Rewon Child, David Luan, Dario Amodei, and Ilya
  Sutskever. 2019.
\newblock Language models are unsupervised multitask learners.

\bibitem[{Ramadan et~al.(2018)Ramadan, Budzianowski, and
  Gasic}]{ramadan2018large}
Osman Ramadan, Pawe{\l} Budzianowski, and Milica Gasic. 2018.
\newblock Large-scale multi-domain belief tracking with knowledge sharing.
\newblock In \emph{Proceedings of the 56th Annual Meeting of the Association
  for Computational Linguistics}, volume~2, pages 432--437.

\bibitem[{Ren et~al.(2019)Ren, Ni, and McAuley}]{ren-etal-2019-scalable}
Liliang Ren, Jianmo Ni, and Julian McAuley. 2019.
\newblock \href {https://doi.org/10.18653/v1/D19-1196} {Scalable and accurate
  dialogue state tracking via hierarchical sequence generation}.
\newblock In \emph{Proceedings of the 2019 Conference on Empirical Methods in
  Natural Language Processing and the 9th International Joint Conference on
  Natural Language Processing (EMNLP-IJCNLP)}, pages 1876--1885, Hong Kong,
  China. Association for Computational Linguistics.

\bibitem[{Serban et~al.(2016)Serban, Sordoni, Bengio, Courville, and
  Pineau}]{serban2016building}
Iulian~V Serban, Alessandro Sordoni, Yoshua Bengio, Aaron Courville, and Joelle
  Pineau. 2016.
\newblock Building end-to-end dialogue systems using generative hierarchical
  neural network models.
\newblock In \emph{Thirtieth AAAI Conference on Artificial Intelligence}.

\bibitem[{Shu et~al.(2019)Shu, Molino, Namazifar, Xu, Liu, Zheng, and
  Tur}]{shu-etal-2019-flexibly}
Lei Shu, Piero Molino, Mahdi Namazifar, Hu~Xu, Bing Liu, Huaixiu Zheng, and
  Gokhan Tur. 2019.
\newblock \href {https://doi.org/10.18653/v1/W19-5922} {Flexibly-structured
  model for task-oriented dialogues}.
\newblock In \emph{Proceedings of the 20th Annual SIGdial Meeting on Discourse
  and Dialogue}, pages 178--187, Stockholm, Sweden. Association for
  Computational Linguistics.

\bibitem[{Srivastava et~al.(2014)Srivastava, Hinton, Krizhevsky, Sutskever, and
  Salakhutdinov}]{srivastava2014dropout}
Nitish Srivastava, Geoffrey Hinton, Alex Krizhevsky, Ilya Sutskever, and Ruslan
  Salakhutdinov. 2014.
\newblock Dropout: a simple way to prevent neural networks from overfitting.
\newblock \emph{The Journal of Machine Learning Research}, 15(1):1929--1958.

\bibitem[{Su et~al.(2018)Su, Lo, Yeh, and Chen}]{su-etal-2018-natural}
Shang-Yu Su, Kai-Ling Lo, Yi-Ting Yeh, and Yun-Nung Chen. 2018.
\newblock \href {https://doi.org/10.18653/v1/N18-2010} {Natural language
  generation by hierarchical decoding with linguistic patterns}.
\newblock In \emph{Proceedings of the 2018 Conference of the North {A}merican
  Chapter of the Association for Computational Linguistics: Human Language
  Technologies, Volume 2 (Short Papers)}, pages 61--66, New Orleans, Louisiana.
  Association for Computational Linguistics.

\bibitem[{Sutskever et~al.(2014)Sutskever, Vinyals, and Le}]{NIPS2014_5346}
Ilya Sutskever, Oriol Vinyals, and Quoc~V Le. 2014.
\newblock \href
  {http://papers.nips.cc/paper/5346-sequence-to-sequence-learning-with-neural-networks.pdf}
  {Sequence to sequence learning with neural networks}.
\newblock In Z.~Ghahramani, M.~Welling, C.~Cortes, N.~D. Lawrence, and K.~Q.
  Weinberger, editors, \emph{Advances in Neural Information Processing Systems
  27}, pages 3104--3112. Curran Associates, Inc.

\bibitem[{Szegedy et~al.(2016)Szegedy, Vanhoucke, Ioffe, Shlens, and
  Wojna}]{szegedy2016rethinking}
Christian Szegedy, Vincent Vanhoucke, Sergey Ioffe, Jon Shlens, and Zbigniew
  Wojna. 2016.
\newblock Rethinking the inception architecture for computer vision.
\newblock In \emph{Proceedings of the IEEE conference on computer vision and
  pattern recognition}, pages 2818--2826.

\bibitem[{Vaswani et~al.(2017)Vaswani, Shazeer, Parmar, Uszkoreit, Jones,
  Gomez, Kaiser, and Polosukhin}]{vaswani17attention}
Ashish Vaswani, Noam Shazeer, Niki Parmar, Jakob Uszkoreit, Llion Jones,
  Aidan~N Gomez, \L~ukasz Kaiser, and Illia Polosukhin. 2017.
\newblock \href
  {http://papers.nips.cc/paper/7181-attention-is-all-you-need.pdf} {Attention
  is all you need}.
\newblock In I.~Guyon, U.~V. Luxburg, S.~Bengio, H.~Wallach, R.~Fergus,
  S.~Vishwanathan, and R.~Garnett, editors, \emph{Advances in Neural
  Information Processing Systems 30}, pages 5998--6008. Curran Associates, Inc.

\bibitem[{Wen et~al.(2016)Wen, Gasic, Mrk{\v{s}}i{\'c}, Rojas~Barahona, Su,
  Ultes, Vandyke, and Young}]{wen2016conditional}
Tsung-Hsien Wen, Milica Gasic, Nikola Mrk{\v{s}}i{\'c}, Lina~M. Rojas~Barahona,
  Pei-Hao Su, Stefan Ultes, David Vandyke, and Steve Young. 2016.
\newblock \href {https://doi.org/10.18653/v1/D16-1233} {Conditional generation
  and snapshot learning in neural dialogue systems}.
\newblock In \emph{Proceedings of the 2016 Conference on Empirical Methods in
  Natural Language Processing}, pages 2153--2162, Austin, Texas. Association
  for Computational Linguistics.

\bibitem[{Wen et~al.(2015)Wen, Ga{\v{s}}i{\'c}, Mrk{\v{s}}i{\'c}, Su, Vandyke,
  and Young}]{wen-etal-2015-semantically}
Tsung-Hsien Wen, Milica Ga{\v{s}}i{\'c}, Nikola Mrk{\v{s}}i{\'c}, Pei-Hao Su,
  David Vandyke, and Steve Young. 2015.
\newblock \href {https://doi.org/10.18653/v1/D15-1199} {Semantically
  conditioned {LSTM}-based natural language generation for spoken dialogue
  systems}.
\newblock In \emph{Proceedings of the 2015 Conference on Empirical Methods in
  Natural Language Processing}, pages 1711--1721, Lisbon, Portugal. Association
  for Computational Linguistics.

\bibitem[{Wen et~al.(2017)Wen, Vandyke, Mrk{\v{s}}i{\'c}, Gasic,
  Rojas~Barahona, Su, Ultes, and Young}]{wen2016network}
Tsung-Hsien Wen, David Vandyke, Nikola Mrk{\v{s}}i{\'c}, Milica Gasic, Lina~M.
  Rojas~Barahona, Pei-Hao Su, Stefan Ultes, and Steve Young. 2017.
\newblock \href {https://www.aclweb.org/anthology/E17-1042} {A network-based
  end-to-end trainable task-oriented dialogue system}.
\newblock In \emph{Proceedings of the 15th Conference of the {E}uropean Chapter
  of the Association for Computational Linguistics: Volume 1, Long Papers},
  pages 438--449, Valencia, Spain. Association for Computational Linguistics.

\bibitem[{Williams et~al.(2017)Williams, Asadi, and
  Zweig}]{williams-etal-2017-hybrid}
Jason~D. Williams, Kavosh Asadi, and Geoffrey Zweig. 2017.
\newblock \href {https://doi.org/10.18653/v1/P17-1062} {Hybrid code networks:
  practical and efficient end-to-end dialog control with supervised and
  reinforcement learning}.
\newblock In \emph{Proceedings of the 55th Annual Meeting of the Association
  for Computational Linguistics (Volume 1: Long Papers)}, pages 665--677,
  Vancouver, Canada. Association for Computational Linguistics.

\bibitem[{Wu et~al.(2019{\natexlab{a}})Wu, Madotto, Hosseini-Asl, Xiong,
  Socher, and Fung}]{wu-etal-2019-transferable}
Chien-Sheng Wu, Andrea Madotto, Ehsan Hosseini-Asl, Caiming Xiong, Richard
  Socher, and Pascale Fung. 2019{\natexlab{a}}.
\newblock \href {https://doi.org/10.18653/v1/P19-1078} {Transferable
  multi-domain state generator for task-oriented dialogue systems}.
\newblock In \emph{Proceedings of the 57th Annual Meeting of the Association
  for Computational Linguistics}, pages 808--819, Florence, Italy. Association
  for Computational Linguistics.

\bibitem[{Wu et~al.(2019{\natexlab{b}})Wu, Socher, and Xiong}]{wu2019global}
Chien-Sheng Wu, Richard Socher, and Caiming Xiong. 2019{\natexlab{b}}.
\newblock Global-to-local memory pointer networks for task-oriented dialogue.
\newblock In \emph{Proceedings of the International Conference on Learning
  Representations (ICLR)}.

\bibitem[{Yavuz et~al.(2019)Yavuz, Rastogi, Chao, and
  Hakkani-Tur}]{yavuz-etal-2019-deepcopy}
Semih Yavuz, Abhinav Rastogi, Guan-Lin Chao, and Dilek Hakkani-Tur. 2019.
\newblock \href {https://doi.org/10.18653/v1/W19-5917} {{D}eep{C}opy: Grounded
  response generation with hierarchical pointer networks}.
\newblock In \emph{Proceedings of the 20th Annual SIGdial Meeting on Discourse
  and Dialogue}, pages 122--132, Stockholm, Sweden. Association for
  Computational Linguistics.

\bibitem[{Zhang et~al.(2019)Zhang, Ou, and Yu}]{zhang2019task}
Yichi Zhang, Zhijian Ou, and Zhou Yu. 2019.
\newblock Task-oriented dialog systems that consider multiple appropriate
  responses under the same context.
\newblock \emph{arXiv preprint arXiv:1911.10484}.

\bibitem[{Zhao et~al.(2017)Zhao, Lu, Lee, and
  Eskenazi}]{zhao-etal-2017-generative}
Tiancheng Zhao, Allen Lu, Kyusong Lee, and Maxine Eskenazi. 2017.
\newblock \href {https://doi.org/10.18653/v1/W17-5505} {Generative
  encoder-decoder models for task-oriented spoken dialog systems with chatting
  capability}.
\newblock In \emph{Proceedings of the 18th Annual {SIG}dial Meeting on
  Discourse and Dialogue}, pages 27--36, Saarbr{\"u}cken, Germany. Association
  for Computational Linguistics.

\bibitem[{Zhao et~al.(2019)Zhao, Xie, and Eskenazi}]{zhao-etal-2019-rethinking}
Tiancheng Zhao, Kaige Xie, and Maxine Eskenazi. 2019.
\newblock \href {https://doi.org/10.18653/v1/N19-1123} {Rethinking action
  spaces for reinforcement learning in end-to-end dialog agents with latent
  variable models}.
\newblock In \emph{Proceedings of the 2019 Conference of the North {A}merican
  Chapter of the Association for Computational Linguistics: Human Language
  Technologies, Volume 1 (Long and Short Papers)}, pages 1208--1218,
  Minneapolis, Minnesota. Association for Computational Linguistics.

\bibitem[{Zhong et~al.(2018)Zhong, Xiong, and Socher}]{zhong2018global}
Victor Zhong, Caiming Xiong, and Richard Socher. 2018.
\newblock Global-locally self-attentive encoder for dialogue state tracking.
\newblock In \emph{ACL}.

\bibitem[{Zhou and Small(2019)}]{zhou2019multi}
Li~Zhou and Kevin Small. 2019.
\newblock Multi-domain dialogue state tracking as dynamic knowledge graph
  enhanced question answering.
\newblock \emph{arXiv preprint arXiv:1911.06192}.

\end{thebibliography}
\bibliographystyle{acl_natbib}

\appendix

\clearpage
\pagebreak 

\section{Data Pre-processing}
\label{app:data}

First, we delexicalize each target system response sequence by replacing the matched entity attribute that appears in the sequence to the canonical tag $\langle domain\_slot \rangle$. For example, the original target response `the train id is tr8259 departing from cambridge' is delexicalized into `the train id is \textit{train\_id} departing from \textit{train\_departure}'. We use the provided entity databases (DBs) to match potential attributes in all target system responses. 
To construct dialogue history, we keep the original version of all text, including system responses of previous turns, rather than the delexicalized form.
We split all sequences of dialogue history, user utterances of the current turn, dialogue states, and delexicalized target responses, into case-insensitive tokens.
We share the embedding weights of all source sequences, including dialogue history, user utterance, and dialogue states, but use a separate embedding matrix to encode the target system responses.

\noindent We summarize the number of dialogues in each domain in Table \ref{tab:domain_data}. For each domain, a dialogue is selected as long as the whole dialogue (i.e. single-domain dialogue) or parts of the dialogue (i.e. in multi-domain dialogue) is involved with the domain. For each domain, we also build a set of possible \textit{inform} and \textit{request} slots using the dialogue state annotation in the training data. The details of slots and database in each domain can be seen in Table \ref{tab:db}. The DBs of 3 domains \textit{taxi, police}, and \textit{hospital} are not available as part of the benchmark.
On average, each dialogue has 1.8 domains and extends over 13 turns. 

%In this work, we do not track booking-related slots such as \textit{booking\_people} or \textit{booking\_day} in each domain because the knowledge base for booking was not provided in the benchmark. 
%We assume that not tracking booking-related slots should not affect the performance of task completion (as there is no knowledge base to determine booking constraints). We leave this part for further extension in the future work.

\begin{table*}[htbp]
\centering
\small
\resizebox{1.0\textwidth}{!} {
\begin{tabular}{lp{6cm}lp{5cm}}
\hline
\multicolumn{1}{c}{\textbf{Domain}} & \multicolumn{1}{c}{\textbf{Slots}}                                                                                                                                                                & \multicolumn{1}{c}{\textbf{\#entities}} & \multicolumn{1}{c}{\textbf{DB attributes}}                                                                                          \\ \hline
\textbf{Restaurant}                 & inf\_area, inf\_food, inf\_name, inf\_pricerange, 
inf\_bookday, inf\_bookpeople, inf\_booktime, 
req\_address, req\_area, req\_food, req\_phone, req\_postcode                                                                                   & 110                                     & id, address, area, food, introduction, name, phone, postcode, pricerange, signature, type                                           \\ \hline
\textbf{Hotel}                      & inf\_area, inf\_internet, inf\_name, inf\_parking, inf\_pricerange, inf\_stars, inf\_type, 
inf\_bookday, inf\_bookpeople, inf\_bookstay, 
req\_address, req\_area, req\_internet, req\_parking, req\_phone, req\_postcode, req\_stars, req\_type & 33                                      & id,  address,  area,  internet,  parking,  single,  double,  family, name, phone, postcode, pricerange’, takesbookings, stars, type \\ \hline
\textbf{Attraction}                 & inf\_area, inf\_name, inf\_type, req\_address, req\_area, req\_phone, req\_postcode, req\_type                                                                                                    & 79                                      & id, address, area, entrance, name, phone, postcode, pricerange, openhours, type                                                     \\ \hline
\textbf{Train}                      & inf\_arriveBy, inform\_day, inf\_departure, inf\_destination, inf\_leaveAt, 
inf\_bookpeople,
req\_duration, req\_price                                                                                             & 2,828                                   & trainID, arriveBy, day, departure, destination, duration, leaveAt, price                                                            \\ \hline
\textbf{Taxi}                       & inf\_arriveBy, inf\_departure, inf\_destination, inf\_leaveAt, req\_phone                                                                                                                         & \textbf{-}                              & \textbf{-}                                                                                                                          \\ \hline
\textbf{Police}                     & inf\_department, req\_address, req\_phone, req\_postcode                                                                                                                                          & \textbf{-}                              & \textbf{-}                                                                                                                          \\ \hline
\textbf{Hospital}                   & req\_address, req\_phone, req\_postcode                                                                                                                                                           & -                                       & -                                                                                                                                   \\ \hline
\end{tabular}
}
\caption{Summary of slots and DB details by domain in the MultiWOZ dataset \citep{budzianowski2018multiwoz}}
\label{tab:db}
\end{table*}

\section{Baselines}
\label{app:baselines}
We describe our baseline models in DST, context-to-text generation, and end-to-end dialogue tasks. 

\subsection{DST}
\label{app:dst_baselines}
\textbf{FJST} and \textbf{HJST} \citep{eric2019multiwoz}.
These models adopt a fixed-vocabulary DST approach. Both models include encoder modules (either bidirectional LSTM or hierarchical LSTM) to encode the dialogue history. 
The models pass the context hidden states to separate linear transformation to obtain final vectors to predict individual slots separately. 
The output vector is used to measure a score of each candidate from a predefined candidate set. 

\noindent \textbf{DST Reader} \citep{gao2019dialog}. 
This model considers the DST task as a reading comprehension task and predicts each slot as a span over tokens within dialogue history. 
DST Reader utilizes attention-based neural networks with additional modules to predict slot type and carryover probability.

\noindent \textbf{TSCP} \citep{lei2018sequicity}. 
%TSCP is an end-to-end dialogue system that can do both DST and NLG. 
The model adopts a sequence-to-sequence framework with a pointer network to generate dialogue states. 
The source sequence is a combination of the last user utterance, dialogue state of the previous turn, and user utterance.
To compare with TSCP in a multi-domain task-oriented dialogue setting, we adapt the model to multi-domain dialogues by formulating the dialogue state of the previous turn similarly as our models.
We reported the performance when the maximum length of the output dialogue state sequence $L$ is set to 20 tokens (original default parameter is 8 tokens but we expect longer dialogue state in MultiWOZ benchmark and selected 20 tokens). 

\noindent \textbf{HyST} \citep{goel2019hyst}. 
This model combines the advantage of fixed-vocabulary and open-vocabulary approaches. 
The model uses an open-vocabulary approach in which the set of candidates of each slot is constructed based on all word n-grams in the dialogue history. 
Both approaches are applied in all slots and depending on their performance in the validation set, the better approach is used to predict individual slots during test time.

\noindent \textbf{TRADE} \citep{wu-etal-2019-transferable}. 
The model adopts a sequence-to-sequence framework with a pointer network to generate individual slot token-by-token. 
The prediction is additionally supported by a slot gating component that decides whether the slot is ``none", ``dontcare", or ``generate". When the gate of a slot is predicted as ``generate", the model will generate value as a natural output sequence for that slot.

\noindent \textbf{NADST} \cite{Le2020Non-Autoregressive}.
The model proposes a non-autoregressive approach for dialogue state tracking which enables learning dependencies between domain-level and slot-level representations as well as token-level representations of slot values. 

\noindent \textbf{DSTQA} \cite{zhou2019multi}.
The model treats dialogue state tracking as a question answering problem in which state values can be predicted through lexical spans or unique generated values. It is enhanced with a knowledge graph where each node represent a slot and edges are based on overlaps of their value sets. 

\noindent \textbf{SOM-DST} \cite{kim-etal-2020-efficient}.
This is the current state-of-the-art model on the MultiWOZ2.1 dataset. 
The model exploits a selectively overwriting mechanism on a fixed-sized memory of dialogue states. At each dialogue turn, the mechanism involve decision making on whether to update or carryover the state values from previous turns. 

\subsection{Context-to-Text Generation}
\noindent \textbf{Baseline}. \cite{budzianowski2018multiwoz} provides a baseline for this setting by following the sequence-to-sequence model \citep{NIPS2014_5346}.
The source sequence is all past dialogue turns and the target sequence is the system response. 
The initial hidden state of the RNN decoder is incorporated with additional signals from the dialogue states and database representations. 

\noindent \textbf{TokenMoE} \citep{pei2019modular}. TokenMoE refers to Token-level Mixture-of-Expert model. The model follows a modularized approach by separating different components known as expert bots for different dialogue scenarios. A dialogue scenario can be dependent on a domain, a type of dialogue act, etc. A chair bot is responsible for controlling expert bots to dynamically generate dialogue responses. 

\noindent \textbf{HDSA} \citep{chen2019semantically}. This is the current state-of-the-art in terms of Inform and BLEU score in the context-to-text generation setting in MultiWOZ2.0.
HDSA leverages the structure of dialogue acts to build a multi-layer hierarchical graph. The graph is incorporated as an inductive bias in a self-attention network to improve the semantic quality of generated dialogue responses. 

\noindent \textbf{Structured Fusion} \citep{mehri2019structured}. This approach follows a traditional modularized dialogue system architecture, including separate components for NLU, DM, and NLG. These components are pre-trained and combined into an end-to-end system. Each component output is used as a structured input to other components. 

\noindent \textbf{LaRL} \citep{zhao-etal-2019-rethinking}. This model uses a latent dialogue action framework instead of handcrafted dialogue acts. The latent variables are learned using unsupervised learning with stochastic variational inference. The model is trained in a reinforcement learning framework whereby the parameters are trained to yield a better Success rate. 
The model is the current state-of-the-art in terms of Success metric. 

\noindent \textbf{GPT2} \cite{budzianowski-vulic-2019-hello}. 
Unsupervised pre-training language models have significantly improved machine learning performance in many NLP tasks. 
This baseline model leverages the power of a pre-trained model \cite{radford2019language} and adapts to the context-to-text generation setting in task-oriented dialogues. 
All input components, including dialogue state and database state, are transformed into raw text format and concatenated as a single sequence. The sequence is used as input to a pre-trained GPT-2 model which is then fine-tuned with MultiWOZ data.

\noindent \textbf{DAMD} \cite{zhang2019task}.
This is the current state-of-the-art model for context-to-text generation task in MultiWOZ 2.1. This approach augments training data with multiple responses of similar context. Each dialogue state is mapped to multiple valid dialogue acts to create additional state-act pairs. 

\subsection{End-to-End}
\noindent \textbf{TSCP} \citep{lei2018sequicity}. 
In addition to the DST task, we evaluate TSCP as an end-to-end dialogue system that can do both DST and NLG. 
We adapt the models to the multi-domain DST setting as described in Section \ref{app:dst_baselines} and keep the original response decoder. 
Similar to the DST component, the response generator of TSCP also adopts a pointer network to generate tokens of the target system responses by copying tokens from source sequences. 
In this setting, we test TSCP with two settings of the maximum length of the output dialogue state sequence: $L=8$ and $L=20$. 

\noindent \textbf{HRED-TS} \cite{peng2019teacher}. 
This model adopts a teacher-student framework to address multi-domain task-oriented dialogues.
Multiple teacher networks are trained for different domains and intermediate representations of dialogue acts and output responses are used to guide a universal student network. 
The student network uses these representations to directly generate responses from dialogue context without predicting dialogue states. 

\section{Qualitative Analysis}
\label{app:qual_res}
\begin{figure*}[htbp]
    \centering
	\includegraphics[width=1\columnwidth]{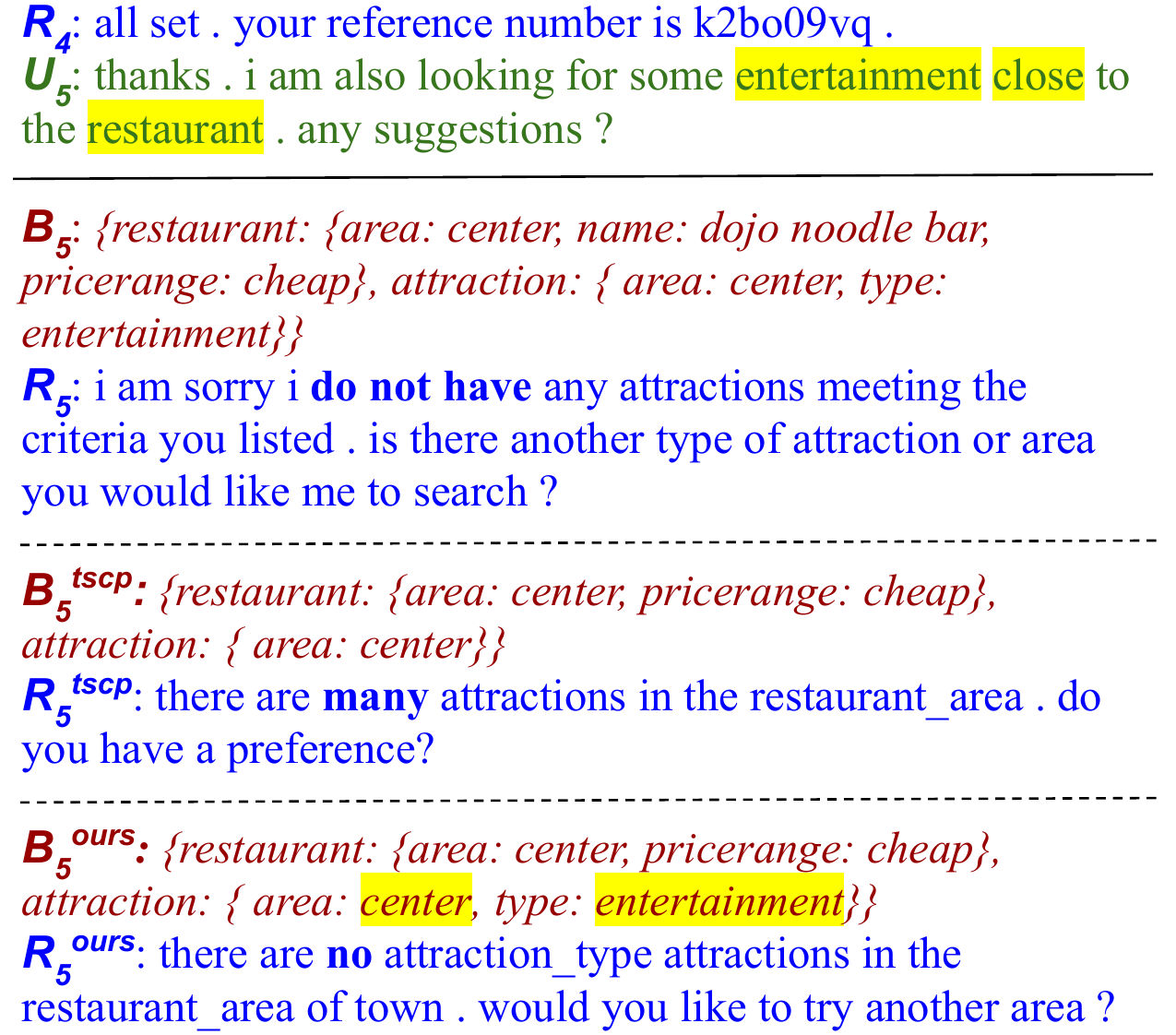}
	\caption{Example dialogue with the input system response $R_{t-1}$ and current user utterance $U_t$, and the output state $BS_t$ and system response $R_t$. Compared with TSCP, our dialogue state and response are more correct and closer to the ground truth.}
	%Visualization of attention to the user utterance sequence at slot-level (lower right) and domain-level (upper right) is also included. More red denotes higher attention score between domain or slot representation and token representation. Best viewed in color.}
	\label{fig:visual}
\end{figure*}

\begin{figure*}[htbp]
    \centering
	\includegraphics[width=1\textwidth]{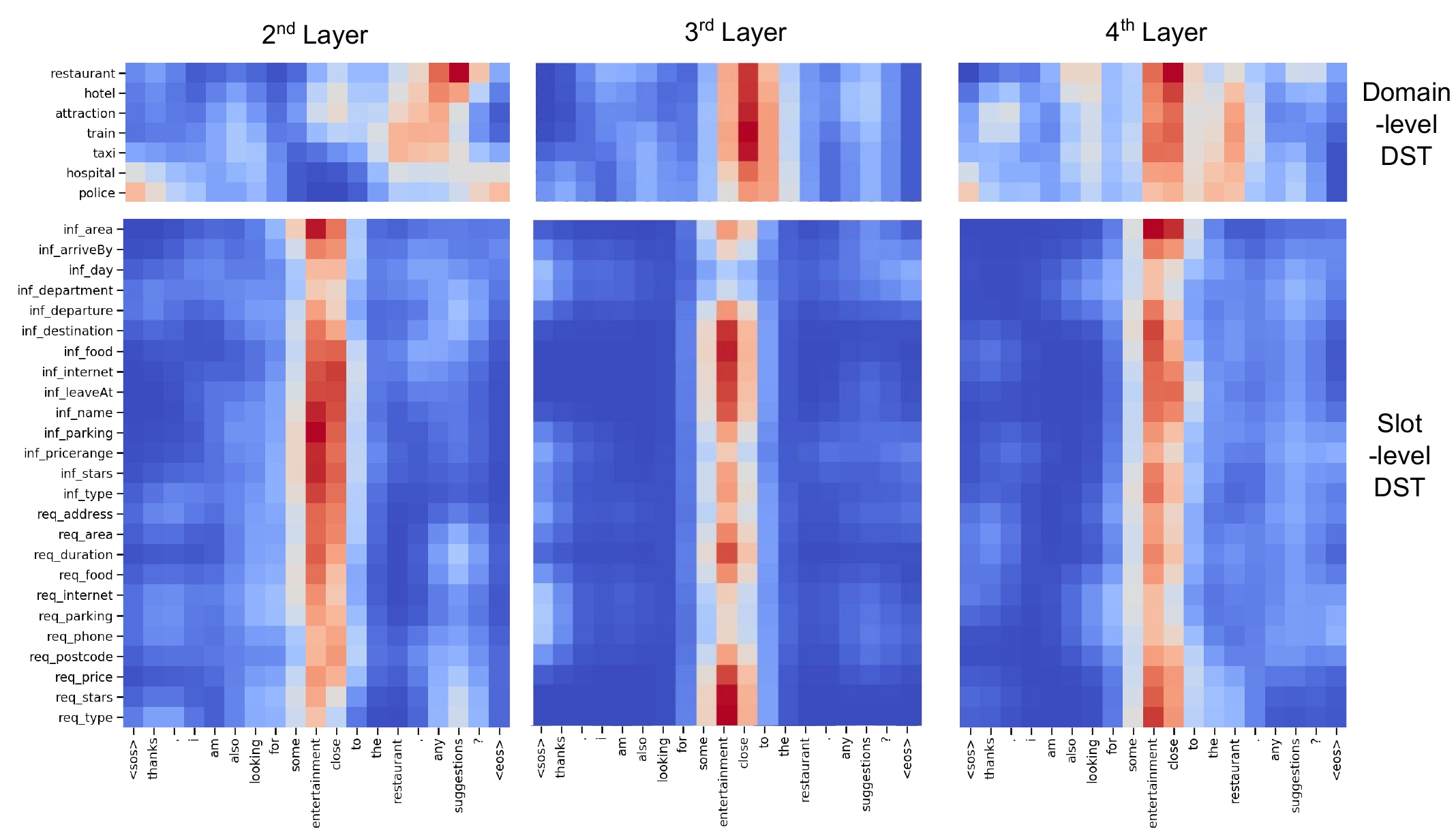}
	\caption{
	Visualization of attention to the user utterance sequence at slot-level (lower right) and domain-level (upper right) is also included. More red denotes higher attention score between domain or slot representation and token representation. Best viewed in color.}
	\label{fig:visual_attn}
\end{figure*}

We examine an example of dialogue in the test data and compare our predicted outputs with the baseline TSCP ($L=20$) \citep{lei2018sequicity} and the ground truth. 
From Figure \ref{fig:visual}, we observe that both our predicted dialogue state and system response are more correct than the baseline. Specifically, our dialogue state can detect the correct \textit{type} slot in the \textit{attraction} domain. As our dialogue state is correctly predicted, the queried results from DB is also more correct, resulting in better response with the right information (i.e. `no attraction available'). 
In Figure \ref{fig:visual_attn}, we show the visualization of domain-level and slot-level attention on the user utterance.
We notice important tokens of the text sequences, i.e. `entertainment' and `close to', are attended with higher attention scores. Besides, at domain-level attention, we find a potential additional signal from the token `restaurant', which is also the domain from the previous dialogue turn. We also observe that attention is more refined throughout the neural network layers. For example, in the domain-level processing, compared to the $2^{nd}$ layer, the $4^{th}$ layer attention is more clustered around specific tokens of the user utterance. 
%The complete predicted output for this example dialogue and other qualitative analysis can be seen in Table \ref{tab:sample_dial}.

In Table \ref{tab:sample_dial} and \ref{tab:sample_dial_cont}, we reported the complete output of this example dialogue. Overall, our dialogue agent can carry a proper dialogue with the user throughout the dialogue steps.
%The dialogue extends across 3 domains: \textit{restaurant, attraction}, and \textit{taxi} sequentially. 
Specifically, we observed that our model can detect new domains at dialogue steps where the domains are introduced e.g. \textit{attraction} domain at the $5^{th}$ turn and \textit{taxi} domain at the $8^{th}$ turn. 
The dialogue agent can also detect some of the co-references among the domains. For example, at the $5^{th}$ turn, the dialogue agent can infer the slot \textit{area} for the new domain \textit{attraction} as the user mentioned `close the restaurant'. 
We noticed that that at later dialogue steps such as the $6^{th}$ turn, our decoded dialogue state is not correct possibly due to the incorrect decoded dialogue state in the previous turn, i.e. $5^{th}$ turn. 
%We also noted that at the $3^{rd}$ turn, our predicted response is different from the ground-truth response because we do not perform state tracking for booking-related slots.

\begin{figure}[htbp]
    \centering
	\includegraphics[width=1\columnwidth]{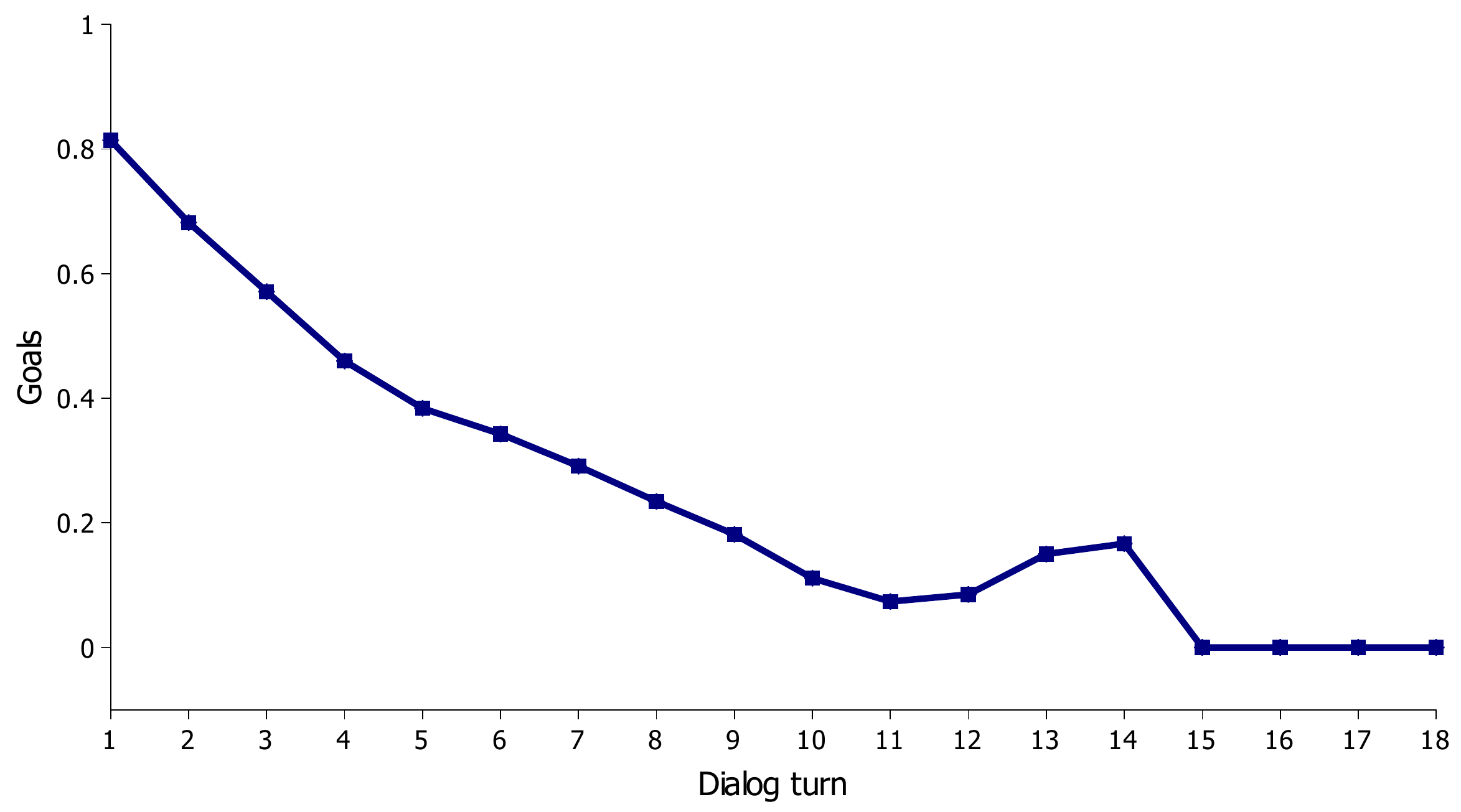}
	\caption{Joint Accuracy metric by dialogue turn in the test data.}
	\label{fig:turn_goal}
\end{figure}

\begin{figure}[htbp]
    \centering
	\includegraphics[width=1\columnwidth]{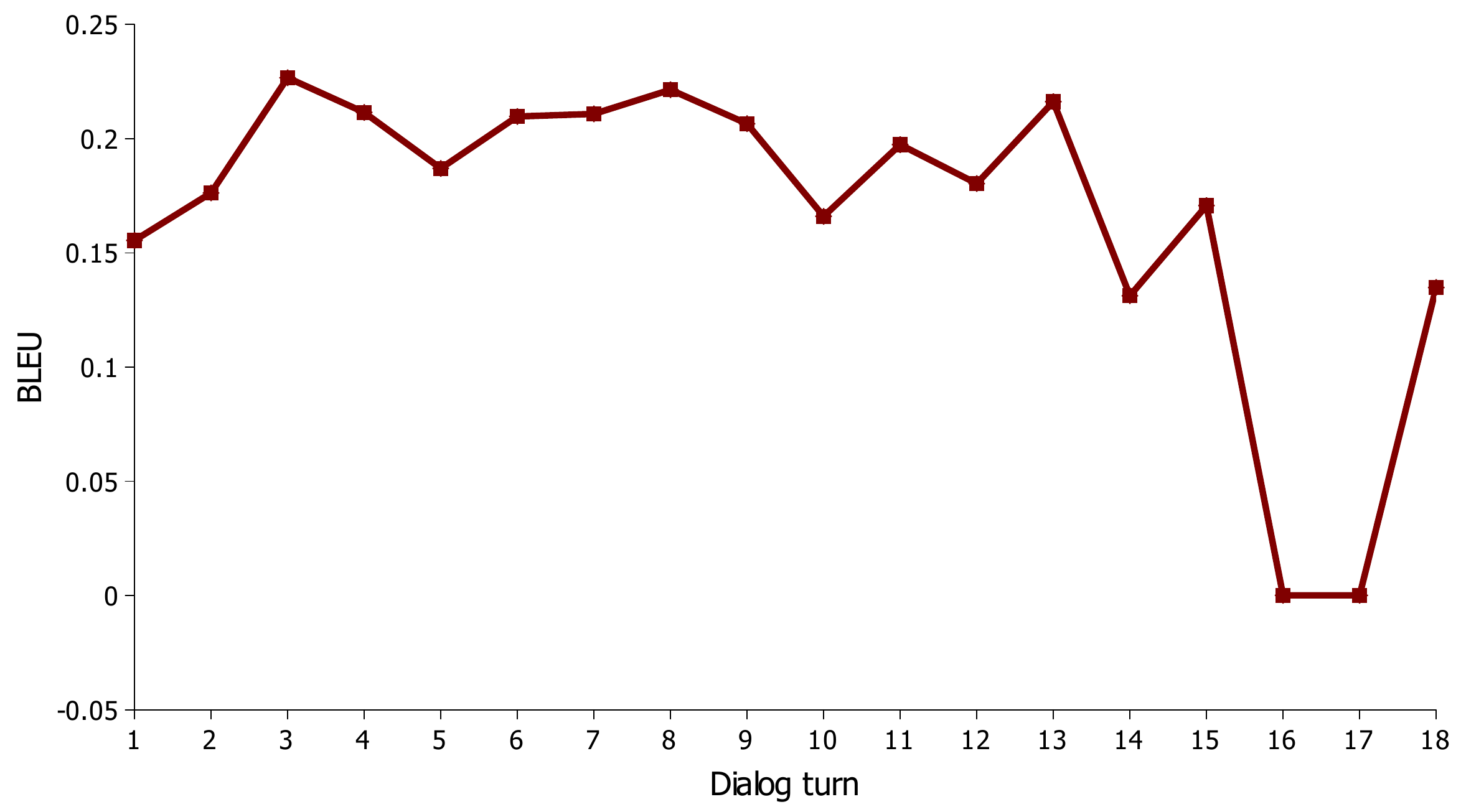}
	\caption{BLEU4 metric by dialogue turn in the test data.}
	\label{fig:turn_bleu}
	%\vspace{-0.6cm}
\end{figure}

In Figure \ref{fig:turn_goal} and \ref{fig:turn_bleu}, we plotted the Joint Goal Accuracy and BLEU metrics of our model by dialogue turn. As we expected, the Joint Accuracy metric tends to decrease as the dialogue history extends over time. 
The dialogue agent achieves the highest accuracy in state tracking at the $1^{st}$ turn and gradually reduces to zero accuracy at later dialogue steps, i.e. $15^{th}$ to $18^{th}$ turns. For response generation performance, the trend of BLEU score is less obvious. The dialogue agent obtains the highest BLEU scores at the $3^{rd}$ turn and fluctuates between the $2^{nd}$ and $13^{th}$ turn. 
%The inconsistency of BLEU performance could be explained by the delexicalization procedure of system responses i.e. the dialogue agent can output a correct delexicalized response even in cases of wrongly detected dialogue belief states. 

\begin{table*}[htbp]
\centering
\small
\resizebox{1.0\textwidth}{!} {
\begin{tabular}{p{0.05cm}p{4cm}p{5.1cm}p{5.1cm}}
\hline
\multicolumn{1}{c}{\textbf{Turn}}  & \multicolumn{1}{c}{\textbf{Input Dialogue}}                                                                                                                    & \multicolumn{1}{c}{\textbf{Ground-truth State and Response}}                                                                                                                                                    & \multicolumn{1}{c}{\textbf{Our Predicted State and Response}}                                                                                  \\ \hline
& $R_0$: \textless{}blank\textgreater{}                                                                                                                       & $B_1$: \{restaurant: \{area: centre, pricerange: cheap\}\}                                                                                                                                                                 & $B_1$: \{restaurant: \{area: centre, pricerange: cheap\}\}                                                                                            \\ 
1 & $U_1$: i am looking for a cheap restaurant in the center of the city                                                                                        & $R_1$: do you have any specific type of food you would like ?                                                                                                                                                               & $R_1$: there are 15 restaurant\_pricerange restaurant\_type s in the restaurant\_area of town . do you have a preference ?                             \\ \hline
 & $U_1$: do you have any specific type of food you would like ?                                                                                               & $B_2$: \{restaurant: \{area: centre, pricerange: cheap\}\}                                                                                                                                                                 & $B_2$: \{restaurant: \{area: centre, pricerange: cheap\}\}                                                                                            \\ 
2 & $U_2$: no , i am not picky as long as the price -s are low .                                                                                                & $R_2$: there is a restaurant\_pricerange restaurant\_food restaurant\_type called the restaurant\_name located in the restaurant\_area of town . would you like to book a table ?                                           & $R_2$: how about restaurant\_name ? it s an restaurant\_pricerange restaurant\_food restaurant\_type located in the restaurant\_area of town .         \\ \hline
& $R_2$: there is a cheap chinese restaurant called the dojo noodle bar located in the centre of town . would you like to book a table ?                      & $B_3$: \{restaurant: \{area: centre, pricerange: cheap,  name: dojo noodle bar\}\}                                                                                                                                         & $B_2$: \{restaurant: \{area: centre, pricerange: cheap\}\}                                                                                            \\ 
3 & $U_3$: yes please , for 8 people at 1830 on thursday .                                                                                                      & $R_3$: i am sorry but restaurant\_name is solidly booked at that time . i can try a different time or day for you .                                                                                                         & $R_3$: booking was successful . the table will be reserved for 15 minutes . reference number is restaurant\_reference .                                \\ \hline
& $R_3$: i am sorry but dojo noodle bar is solidly booked at that time . i can try a different time or day for you                                            & $B_4$: \{restaurant: \{area: centre, pricerange: cheap,  name: dojo noodle bar\}\}                                                                                                                                         & $B_4$: \{restaurant: \{area: centre, pricerange: cheap\}\}                                                                                            \\ 
4 & $U_4$: can you try to book it at 1730 .                                                                                                                     & $R_4$: all set . your reference number is restaurant\_reference .                                                                                                                                                           & $R_4$: booking was successful . the table will be reserved for 15 minutes . reference number is restaurant\_reference .                                \\ \hline
& $R_4$: all set . your reference number is k2bo09vq .                                                                                                        & $B_5$: \{restaurant: \{area: centre, pricerange: cheap,  name: dojo noodle bar\}, attraction: \{area: centre, type: entertainment\}\}                                                                                      & $B_5$: \{restaurant: \{area: centre, pricerange: cheap\}, attraction: \{area: centre, type: entertainment\}\}                                         \\ 
5 & $U_5$: thanks . i am also looking for some entertainment close to the restaurant . any suggestions ?                                                        & $R_5$: i am sorry i do not have any attractions meeting the criteria you listed . is there another type of attraction or area you would like me to search ?                                                                 & $R_5$: there are no attraction\_type attractions in the restaurant\_area of town . would you like to try another area ?                                \\ \hline
& $R_5$: i am sorry i do not have any attractions meeting the criteria you listed . is there another type of attraction or area you would like me to search ? & $B_6$:\{restaurant: \{area: centre, pricerange: cheap,  name: dojo noodle bar\}, attraction: \{area: centre, type: architecture\}\}                                                                                        & $B_6$:\{restaurant: \{area: centre, pricerange: cheap\}, attraction: \{area: centre, type: architecture\}\}                                           \\ 
6 & $U_6$: why do not you try an architectural attraction , instead .                                                                                           & $R_6$: attraction\_name looks good , would you like to head there ?                                                                                                                                                         & $R_6$: attraction\_name is a great attraction\_type in the center of town . would you like  more information ?                                         \\ \hline
& $R_6$: all saints church looks good , would you like to head there ?                                                                                        & $B_7$:\{restaurant: \{area: centre, pricerange: cheap,  name: dojo noodle bar\}, attraction: \{area: centre, type: architecture\}\}                                                                                        & $B_7$:\{restaurant: \{area: centre, pricerange: cheap\}, attraction: \{area: centre, type: architecture\}\}                                           \\ 
7 & $U_7$: yes please . can you let me know what the entrance fee is too ?                                                                                      & $R_7$: its entrance fee is attraction\_pricerange .                                                                                                                                                                         & $R_7$: the entrance fee is attraction\_pricerange . is there anything else i can help you with ?                                                       \\ \hline
& $R_7$: its entrance fee is free .                                                                                                                           & $B_8$:\{restaurant: \{area: centre, pricerange: cheap,  name: dojo noodle bar\}, attraction: \{area: centre, type: architecture\}, taxi: \{departure: dojo noodle bar, destination: all saints church\}\}                  & $B_8$:\{restaurant: \{area: centre, pricerange: cheap\}, attraction: \{area: centre, type: architecture\}, taxi: \{destination: dojo noodle bar\}\}   \\ 
8 & $U_8$: i also need to book a taxi between the restaurant and the church .                                                                                   & $R_8$: what time would you like the taxi to pick you up from restaurant\_name ?                                                                                                                                             & $R_8$: what time would you like to leave or arrive by ?                                                                                                \\ \hline
\end{tabular}
}
\caption{Complete turns of an example multi-domain dialogue with each turn including the input of past system response $R_{t-1}$ and current user utterance $U_t$, and the predicted dialogue state $B_t$ and system response $U_t$. 
The dialogue consists of 11 turns in total and extends across 3 domains sequentially: \textit{restaurant}, \textit{attraction}, and \textit{taxi}. Dialogue continues in Table \ref{tab:sample_dial_cont}.}
\label{tab:sample_dial}
\end{table*}

\begin{table*}[htbp]
\centering
\small
\resizebox{1.0\textwidth}{!} {
\begin{tabular}{p{0.05cm}p{4cm}p{5.1cm}p{5.1cm}}
\hline
\multicolumn{1}{c}{\textbf{Turn}}  & \multicolumn{1}{c}{\textbf{Input Dialogue}}                                                                                                                    & \multicolumn{1}{c}{\textbf{Ground-truth State and Response}}                                                                                                                                                    & \multicolumn{1}{c}{\textbf{Our Predicted State and Response}}                                                                                  \\ \hline
& $R_8$: what time would you like the taxi to pick you up from dojo noodle bar ?                                                                              & $B_9$: \{restaurant: \{area: centre, pricerange: cheap,  name: dojo noodle bar\}, attraction: \{area: centre, type: architecture\}, taxi: \{departure: dojo noodle bar, destination: all saints church, leaveAt: 1830\}\}  & $B_9$:\{restaurant: \{area: centre, pricerange: cheap\}, attraction: \{area: centre, type: architecture\}, taxi: \{destination: dojo noodle bar\}\}   \\ 
9 & $U_9$: 1830 , please .                                                                                                                                      & $R_9$: your taxi is booked . the car is a taxi\_type and the contact number is taxi\_phone . is there anything else i can help you with ?                                                                                   & $R_9$: booking completed ! booked car type taxi\_type contact number taxi\_phone can i help you with anything else ?                                   \\ \hline
& $R_9$: your taxi is booked . the car is a blue bmw and the contact number is 07239644669 . is there anything else i can help you with ?                     & $B_{10}$: \{restaurant: \{area: centre, pricerange: cheap,  name: dojo noodle bar\}, attraction: \{area: centre, type: architecture\}, taxi: \{departure: dojo noodle bar, destination: all saints church, leaveAt: 1830\}\} & $B_{10}$: \{restaurant: \{area: centre, pricerange: cheap\}, attraction: \{area: centre, type: architecture\}, taxi: \{destination: dojo noodle bar\}\} \\ 
10 & $U_{10}$: that s it . thanks .                                                                                                                                & $R_{10}$: awesome , enjoy your trip !                                                                                                                                                                                         & $R_{10}$: you are welcome . have a great day !                                                                                                           \\ \hline
& $R_{10}$: awesome , enjoy your trip !                                                                                                                         & $B_{11}$: \{restaurant: \{area: centre, pricerange: cheap,  name: dojo noodle bar\}, attraction: \{area: centre, type: architecture\}, taxi: \{departure: dojo noodle bar, destination: all saints church, leaveAt: 1830\}\} & $B_{11}$: \{restaurant: \{area: centre, pricerange: cheap\}, attraction: \{area: centre, type: architecture\}, taxi: \{destination: dojo noodle bar\}\} \\ 
11 & $U_{11}$: thanks ! i will definitely try to .                                                                                                                 & $R_{11}$: great ! goodbye and have a nice day .                                                                                                                                                                               & $R_{11}$: thank you for using our services . have a great day .         \\ \hline                                                                                
\end{tabular}
}
\caption{Dialogue continues from Table \ref{tab:sample_dial}. Each turn includes the input of past system response $S_{t-1}$ and current user utterance $U_t$, and the predicted dialogue dialogue state $BS_t$ and system response $S_t$. The dialogue consists of 11 turns in total and extends across 3 domains sequentially: restaurant, attraction, and taxi.}
\label{tab:sample_dial_cont}
\end{table*}

\end{document}